\documentclass{article}

\PassOptionsToPackage{numbers, sort&compress}{natbib}

\usepackage[preprint]{neurips_2019}

\usepackage[utf8]{inputenc} 
\usepackage[T1]{fontenc}    
\usepackage{hyperref}       
\usepackage{url}            
\usepackage{booktabs}       
\usepackage{amsfonts}       
\usepackage{nicefrac}       
\usepackage{microtype}      

\usepackage[pdftex]{graphicx}
\usepackage{enumitem}
\usepackage{subfigure}
\usepackage{wrapfig}
\usepackage{mathtools}
\usepackage{soul}

\usepackage{algorithm}
\usepackage{algpseudocode}

\renewcommand{\vec}[1]{\mathbf{{#1}}}

\title{Hierarchical Autoregressive Image Models\\with Auxiliary Decoders}

\author{
  Jeffrey De Fauw\textsuperscript{*} \quad Sander Dieleman\textsuperscript{*} \quad Karen Simonyan \\
  \textsuperscript{*}Equal contribution \\
  DeepMind, London, UK \\
  \texttt{\{defauw,sedielem,simonyan\}@google.com} \\
}

\begin{document}

\maketitle

\begin{abstract}
Autoregressive generative models of images tend to be biased towards capturing local structure, and as a result they often produce samples which are lacking in terms of large-scale coherence. To address this, we propose two methods to learn discrete representations of images which abstract away local detail. We show that autoregressive models conditioned on these representations can produce high-fidelity reconstructions of images, and that we can train autoregressive priors on these representations that produce samples with large-scale coherence.
We can recursively apply the learning procedure, yielding a hierarchy of progressively more abstract image representations. We train hierarchical class-conditional autoregressive models on the ImageNet dataset and demonstrate that they are able to generate realistic images at resolutions of 128$\times$128 and 256$\times$256 pixels. We also perform a human evaluation study comparing our models with both adversarial and likelihood-based state-of-the-art generative models.
\end{abstract}

\begin{figure}[h]
  \centering
  \includegraphics[width=0.90\linewidth]{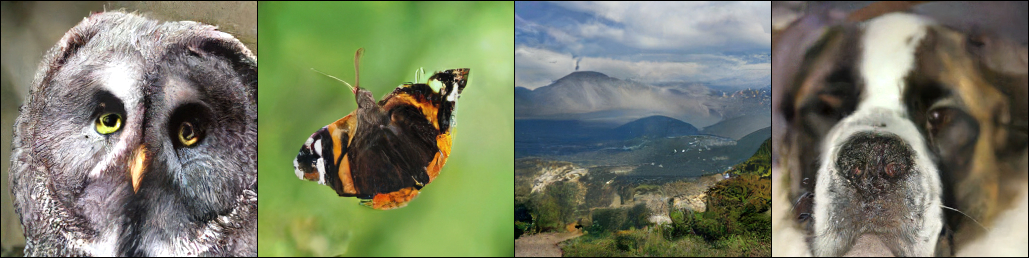}
  \caption{Selected class conditional 256$\times$256 samples from our models. More are available in the appendix and at \url{https://bit.ly/2FJkvhJ}.}
  \label{fig:256_samples}
  \vskip -0.20in
\end{figure}

\section{Introduction}
\label{sec:introduction}
Generative models can be used to model the distribution of natural images. With enough capacity, they are then capable of producing new images from this distribution, which enables the creation of new natural-looking images from scratch. These models can also be conditioned on various annotations associated with the images (e.g. class labels), allowing for some control over the generated output.

In recent years, adversarial learning has proved a powerful tool to create such models~\cite{NIPS2014_5423,radford2015unsupervised,karras2018progressive,brock2018large,nvidiastyle}. An alternative approach is to specify a model in the form of the joint distribution across all pixels, and train the model on a set of images by maximising their likelihood under this distribution (or a lower bound on this likelihood). Several families of models fit into this likelihood-based paradigm, including variational autoencoders (VAEs)~\cite{KingmaW13,pmlr-v32-rezende14}, flow-based models~\cite{DBLP:journals/corr/DinhKB14,45819,NIPS2018_8224} and autoregressive models~\cite{NIPS2015_5637,pmlr-v48-oord16,NIPS2016_6527}.

Likelihood-based models currently lag behind their adversarial counterparts in terms of the visual fidelity and the resolution of their samples. However, adversarial models are known to drop modes of the distribution, something which likelihood-based models are inherently unlikely to do. Within the likelihood-based model paradigm, autoregressive models such as PixelCNN tend to be the best at capturing textures and details in images, because they make no independence assumptions and they are able to use their capacity efficiently through spatial parameter sharing. They also achieve the best likelihoods. We describe PixelCNN in more detail in Section~\ref{sec:background}.

However, autoregressive models are markedly worse at capturing structure at larger scales, and as a result they tend to produce samples that are lacking in terms of large-scale coherence (see appendix for a demonstration). This can be partially attributed to the inductive bias embedded in their architecture, but it is also a consequence of the likelihood loss function, which rewards capturing local correlations much more generously than capturing long-range structure. As far as the human visual system is concerned, the latter is arguably much more important to get right, and this is where adversarial models currently have a substantial advantage.

\begin{wrapfigure}{R}{0.45\textwidth}
    \centering
    \includegraphics[width=0.45\textwidth,trim={6cm 1cm 6cm 1cm},clip]{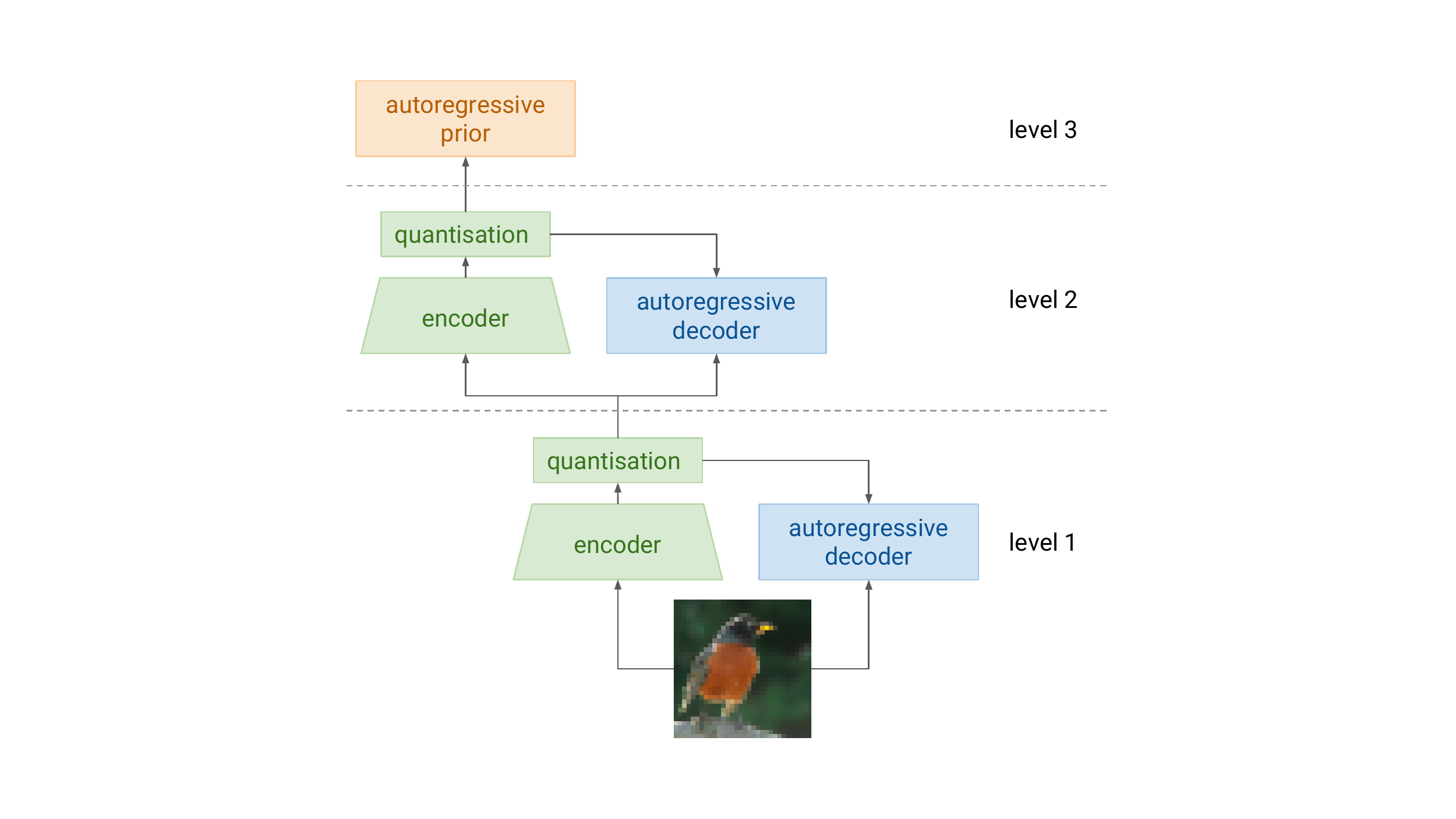}
    \vskip -0.15in
    \caption{Schematic overview of a hierarchical autoregressive model. The dashed lines indicate different stages, which capture different scales of structure in the input image.}
    \label{fig:hierarchy}
    \vskip -0.15in
\end{wrapfigure}

To make autoregressive models pay more attention to large-scale structure, an effective strategy is to remove local detail from the input representation altogether. A simple way to do this for images is by reducing their bit-depth~\cite{NIPS2018_8224,menick2018generating}.
An alternative approach is to learn new input representations that abstract away local detail, by training encoder models. 
We can then train autoregressive models of the image pixels conditioned on these representations, as well as autoregressive priors for these representations, effectively splitting the task into two separate stages~\cite{NIPS2017_7210}. We can extend this approach further by stacking encoder models, yielding a hierarchy of progressively more high-level representations~\cite{NIPS2018_8023}, as shown in Figure~\ref{fig:hierarchy}. This way, we can explicitly assign model capacity to different scales of structure in the images, and turn the bias these models have towards capturing local structure into an advantage. Pseudocode for the full training and sampling procedures is provided in the appendix.

Learning representations that remove local detail while preserving enough information to enable a conditional autoregressive model to produce high-fidelity pixel-level reconstructions is a non-trivial task. A natural way to do this would be to turn the conditional model into an autoencoder, so that the representations and the reconstruction model can be learnt jointly. However, this approach is fraught with problems, as we will discuss in Section~\ref{sec:autoregressive-autoencoders}.

Instead, we propose two alternative strategies based on auxiliary decoders, which are particularly suitable for hierarchical models: we use feed-forward (i.e. non-autoregressive) decoders or \emph{masked self-prediction} (MSP) to train the encoders. Both techniques are described in Section~\ref{sec:auxiliary-decoders}. We show that the produced representations allow us to construct hierarchical models trained using only likelihood losses that successfully produce samples with large-scale coherence. Bringing the capabilities of likelihood-based models up to par with those of their adversarial counterparts in terms of scale and fidelity is important, because this allows us to sidestep any issues stemming from mode dropping and exert more control over the mapping between model capacity and image structure at different scales.

We make the representations learnt by the encoders discrete by inserting vector quantisation (VQ) bottlenecks~\cite{NIPS2017_7210}. This bounds the information content of the representations, and it enables more efficient and stable training of autoregressive priors~\cite{pmlr-v48-oord16}. In addition, models with VQ bottlenecks do not suffer from posterior collapse~\cite{DBLP:conf/conll/BowmanVVDJB16}, unlike regular VAEs. Because of their discrete nature, we will also refer to the learnt representations as \emph{codes}. We cover VQ bottlenecks in neural networks in more detail in Section~\ref{sec:background}. We also include a downsampling operation in the encoders so that higher-level codes have a lower spatial resolution. 

The contributions of this work are threefold: we study the problems associated with end-to-end training of autoencoders with autoregressive decoders. We also propose two alternative strategies for training such models using auxiliary decoders: feed-forward decoding and masked self-prediction (MSP). Finally, we construct hierarchical likelihood-based models that produce high-fidelity and high-resolution samples (128$\times$128 and 256$\times$256) which exhibit large-scale coherence. Selected samples are shown in Figure~\ref{fig:256_samples}.

\section{Background}
\label{sec:background}
We will use PixelCNN as the main building block for hierarchical image models. We will also insert vector quantisation bottlenecks in the encoders to enable them to produce discrete representations. We briefly describe both of these components below and refer to \citet{pmlr-v48-oord16,NIPS2016_6527,NIPS2017_7210} for a more detailed overview.

\subsection{PixelCNN}
PixelCNN is an autoregressive model: it assumes an arbitrary ordering of the pixels and colour channels of an image, and then models the distribution of each intensity value in the resulting sequence conditioned on the previous values. In practice, the intensities are typically flattened into a sequence in `raster scan' order: from top to bottom, then from left to right, and then according to red, green and blue intensities. Let $x_i$ be the intensity value at position $i$ in this sequence. Then the density across all intensity values is factorised into a product of conditionals: $p(\vec{x}) = \prod_{i} p(x_{i}|\vec{x}_{<i})$. PixelCNN models each of these conditionals with the same convolutional neural network, using weight masking to ensure that each value in the sequence depends only on the values before it.

\subsection{Vector quantisation}
Vector quantisation variational autoencoders (VQ-VAE) use a vector quantisation (VQ) bottleneck to learn discrete representations. The encoder produces a continuous $d$-dimensional vector $\vec{z}$, which is then quantised to one of $k$ possible vectors from a codebook. This codebook is learnt jointly with the other model parameters. The quantisation operation is non-differentiable, so gradients are backpropagated through it using straight-through estimation~\cite{bengio2013estimating}. In practice, this means that they are backpropagated into the encoder as if the quantisation operation were absent, which implies that the encoder receives approximate gradients.

The model is trained using the loss function $\mathcal{L} = -\log p(\vec{x} | \vec{z'}) + (\vec{z'} - [\vec{z}])^2 + \beta \cdot ([\vec{z'}] - \vec{z})^2$, where $\vec{x}$ is the input, $\vec{z}$ is the output of the encoder and $\vec{z'}$ is the quantised output. $\beta$ is a hyperparameter and square brackets indicate that the contained expressions are treated as constant w.r.t. differentiation\footnote{$[x]$ is like \texttt{tf.stop\_gradient(x)} in TensorFlow.}. The three terms correspond to the reconstruction log-likelihood, the \emph{codebook loss} and the \emph{commitment loss} respectively. Instead of optimising all terms using gradient descent, we use an alternative learning rule for the codebook using an exponentially smoothed version of K-means, which replaces the codebook loss and which is described in the appendix of \citet{NIPS2017_7210}. Although this approach was introduced in the context of autoencoders, VQ bottlenecks can be inserted in any differentiable model, and we will make use of that fact in Section~\ref{sec:masked-self-prediction-decoders}.

\section{Challenges of autoregressive autoencoding}
\label{sec:autoregressive-autoencoders}
Autoregressive autoencoders are autoencoders with autoregressive decoders. Their appeal lies in the combination of two modelling strategies: using latent variables to capture global structure, and autoregressive modelling to fill in local detail. This idea has been explored extensively in literature~\cite{NIPS2016_6527,gulrajani+al-2016-pixelvae,DBLP:journals/corr/ChenKSDDSSA16,NIPS2017_7210,DBLP:conf/icml/EngelRRDNES17,NIPS2018_8023}.

During training, autoregressive models learn to predict one step ahead given the ground truth. This is often referred to as teacher forcing~\cite{williams:recurrent}. However, when we sample from a trained model, we use previous predictions as the model input instead of ground truth. This leads to a discrepancy between the training and inference procedures: in the latter case, prediction errors can accumulate. 
Unfortunately, autoregressive autoencoders can exhibit several different pathologies, especially when the latent representation has the same spatial structure as the input (i.e., there is a spatial map of latents, not a single latent vector). Most of these stem from an incompatibility between teacher forcing and the autoencoder paradigm:

\begin{wrapfigure}{R}{0.35\textwidth}
    \centering
    \vskip -0.20in
    \includegraphics[width=0.95\linewidth]{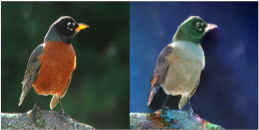}
    \caption{Illustration of an issue with autoregressive autoencoders caused by teacher forcing. Left: original 128$\times$128 image. Right: reconstruction sampled from an autoregressive autoencoder.
    }
    \label{fig:bad_ar_decoder_bird}
    \vskip -0.50in
\end{wrapfigure}

\begin{itemize}[leftmargin=*]
    \item When the loss actively discourages the use of the latent representation to encode information (like the KL term does in VAEs), the decoder will learn to ignore it and use only the autoregressive connections. This phenomenon is known as posterior collapse~\cite{DBLP:conf/conll/BowmanVVDJB16}. 
    
    \item On the other hand, when the latent representation has high information capacity, there is no incentive for the model to learn to use the autoregressive connections in the decoder.
    
    \item The encoder is encouraged to preserve in the latent representations any noise that is present in the input, as the autoregressive decoder cannot accurately predict it from the preceding pixels. This is counter to the intuition that the latent representations should capture high-level information, rather than local noise. This effect is exacerbated by teacher forcing which, during training, enables the decoder to make very good next-step predictions in the absence of noise. When noise is present, there will be a very large incentive for the model to store this information in the codes.
    
    \item The encoder is encouraged to ignore slowly varying aspects of the input that are very predictable from local information, because the ground truth input is always available to the autoregressive decoder during training. This affects colour information in images, for example: it is poorly preserved when sampling image reconstructions, as during inference the sampled intensities will quickly deviate slightly from their original values, which then recursively affects the colour of subsequent pixels. This is demonstrated in Figure~\ref{fig:bad_ar_decoder_bird}.
\end{itemize}

Workarounds to these issues include strongly limiting the capacity of the representation (e.g. by reducing its spatial resolution, using a latent vector without spatial structure, inserting a VQ bottleneck and/or introducing architectural constraints in the encoder) or limiting the receptive field of the decoder~\cite{gulrajani+al-2016-pixelvae,NIPS2017_7210,NIPS2018_8023}. Unfortunately, this limits the flexibility of the models. Instead, we will try to address these issues more directly by decoupling representation learning from autoregressive decoder training.

\section{Auxiliary decoders}
\label{sec:auxiliary-decoders}
To address the issues associated with jointly training encoders and autoregressive decoders, we introduce auxiliary decoders: separate decoder models which are only used to provide a learning signal to the encoders. Once an encoder has been trained this way, we can discard the auxiliary decoder and replace it with a separately trained autoregressive decoder conditioned on the encoder representations. The autoregressive decoder consists of a local model (PixelCNN) and a modulator which maps the encoder representations to a set of biases for each layer of the local model. This separation allows us to design alternative decoders with architectures and loss functions that are tuned for feature learning rather than for reconstruction quality alone.

Although the encoder and decoder are trained using different loss functions, it is still convenient to train them simultaneously, taking care not to backpropagate the autoregressive decoder loss to the encoder. Otherwise, multiple networks would have to be trained in sequence for each level in the hierarchy. We use simultaneous training in all of our experiments. This has a negligible effect on the quality of the reconstructions.

\subsection{Feed-forward decoders}
\label{sec:feedforward-decoders}

The most straightforward form the auxiliary decoder can take is that of a feed-forward model that tries to reconstruct the input. Even though the task of both decoders is then the same, the feed-forward architecture shapes what kinds of information the auxiliary decoder is able to capture. Because such a decoder does not require teacher forcing during training, the issues discussed in Section~\ref{sec:autoregressive-autoencoders} no longer occur.

\begin{figure*}
    \centering
    \subfigure[]{
        \includegraphics[width=0.363\textwidth,trim={6cm 0cm 6cm 0cm},clip]{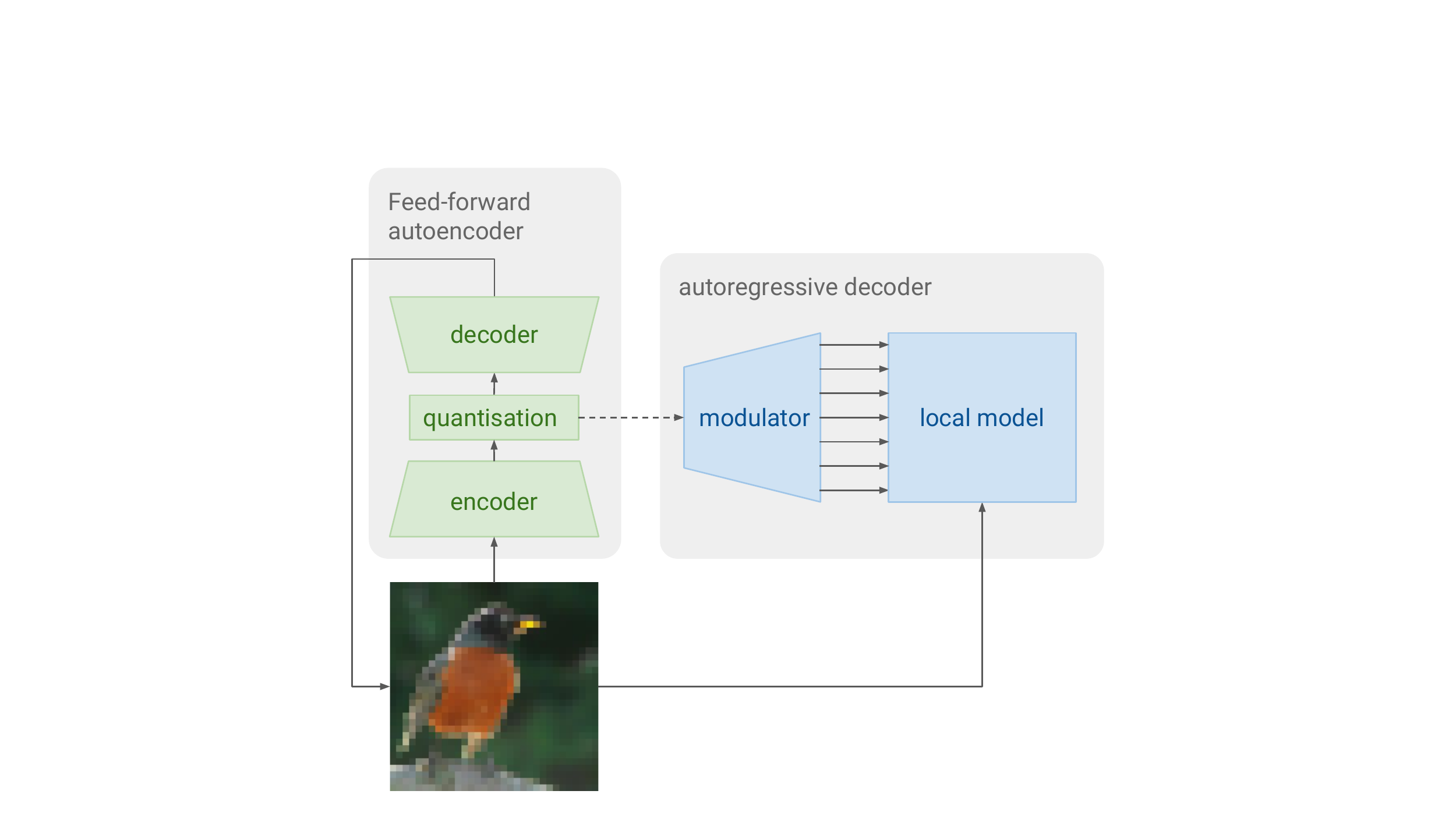} 
        \label{fig:aux-feedforward}
    }
    ~ %
    \subfigure[]{
        \includegraphics[width=0.537\textwidth,trim={3cm 0cm 3cm 0cm},clip]{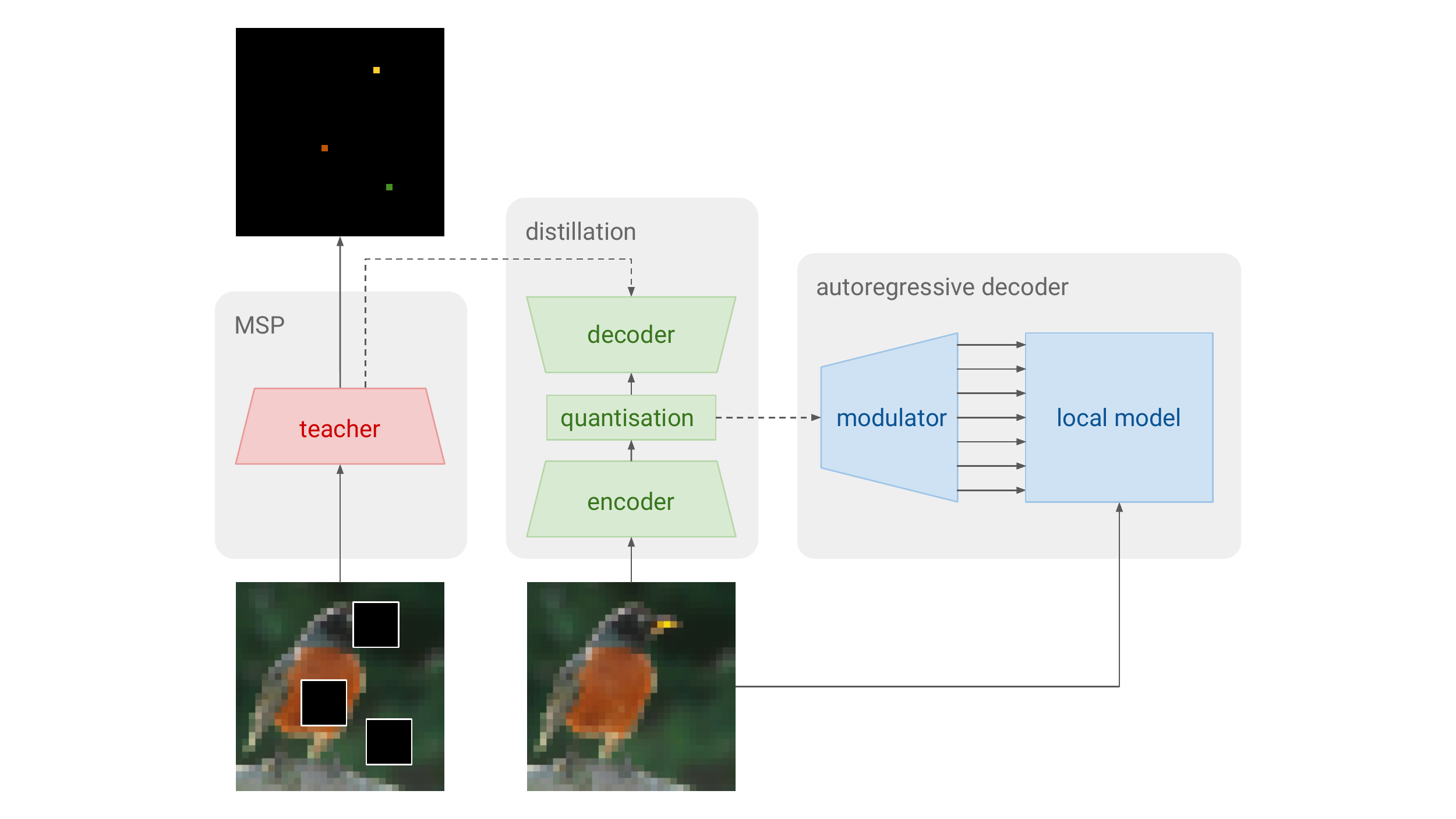} 
        \label{fig:aux-msp}
    }
    \vskip  -0.15in
    \caption{Discrete autoencoders with auxiliary decoders. Left: a feed-forward decoder is used to train the encoder. Right: a masked self-prediction (MSP) model is trained and then distilled into a new model with unmasked input to obtain the encoder. Both models feature autoregressive decoders. Note the dashed arrows indicating that no gradients are backpropagated along these connections.}\label{fig:aux-diagrams}
\end{figure*}

When trained on RGB images, we can treat the pixel intensities as continuous and use the mean squared error (MSE) loss for training. For other types of inputs, such as codes produced by another encoder, we can use the same multinomial log-likelihood that is typically used for autoregressive models. Using the MSE would not make sense as the discrete codes are not ordinal and cannot be treated as continuous values. Figure~\ref{fig:aux-diagrams} (left) shows a diagram of an autoregressive autoencoder with an auxiliary feed-forward decoder.

A significant benefit of this approach is its simplicity: we do not stray too far from the original autoencoder paradigm because the model is still trained using a reconstruction loss. However, an important drawback is that the reconstruction task still encourages the model to capture as much information as possible in the codes, even unimportant details that would be easy for the autoregressive decoder to fill in. It affords relatively little control over the nature and the quantity of information captured in the codes.

\subsection{Masked self-prediction decoders}
\label{sec:masked-self-prediction-decoders}

Models with feed-forward decoders are encouraged to encode as much information as possible in the codes to help the decoders produce detailed reconstructions. As a result, the codes may not be very compressible (see Section~\ref{sec:auxiliary-decoder-design}), which makes stacking multiple encoders to create a hierarchy quite challenging. Instead of optimising the auxiliary decoder for reconstruction, we can train it to do self-prediction: predict the distribution of a pixel given the surrounding pixels. By masking out some region of the input around the pixel to be predicted, we can prevent the decoder from using strong local correlations, and force it to rely on weaker long-range dependencies instead. As a result, the produced codes will be much more compressible because they only contain information about longer-range correlations in the input. The local detail omitted from these representations can later be filled in by the autoregressive decoder.

In practice, \emph{masked self-prediction} (MSP) entails masking some square region of the input and predicting the middle pixel of this region. If the region is 7$\times$7 pixels large, for example, the model can only rely on correlations between pixels that are at least 4 positions away for its prediction. In the presence of noise or other unpredictable local detail (such as textures), the MSP model will be uncertain about its predictions and produce a distribution across all possibilities. The encoder representation will then capture this uncertainty, rather than trying to encode the exact pixel values.

Given an input $\vec{x}$, a pixel position $(i, j)$ and an offset $s$ (corresponding to a mask size of $2s+1$), the MSP objective is to maximise $\log p(x_{ij}|\vec{x} \cdot \vec{m})$, where the input mask $\vec{m}$ is given by:

\begin{equation}
    m_{kl} =  
   \begin{dcases}
     0 \text{ if } i - s \leq k \leq i + s, j - s \leq l \leq j + s \\
     1 \text{ otherwise}
   \end{dcases} .
\end{equation}

In practice, we can select multiple pixel positions per input to make training more sample-efficient, but the total number of positions should be limited because too much of the input could be masked out otherwise.

Because this approach requires input masking, we would have to run a forward pass through the encoder with a different mask for each spatial position if we wanted to compute representations for the entire input. This is computationally prohibitive, so instead, we use distillation~\cite{44873} to obtain an encoder model that does not require its input to be masked. We train a \emph{teacher} model with masked input (a simple feed-forward network), and simultaneously distill its predictions for the selected pixel positions into a \emph{student} model with unmasked input and a vector quantisation bottleneck. Because it is convolutional, this student model will learn to produce valid representations for all spatial positions, even though it is only trained on a subset of spatial positions for each input. Representations for an input can then be computed in a single forward pass. The full setup is visualised in Figure~\ref{fig:aux-diagrams} (right) and described in pseudocode in the appendix.

\section{Related work}
\label{sec:related-work}
Recent work on scaling generative models of images to larger resolutions has been focused chiefly on adversarial models. \citet{karras2018progressive,nvidiastyle} trained generative adversarial networks (GANs) that can generate various scenes and human faces at resolutions of 256$\times$256 and higher (up to 1 megapixel for the latter). \citet{brock2018large} generate 512$\times$512 images for each of the 1000 classes in the ImageNet dataset, all with the same GAN model.

\citet{DBLP:conf/icml/ReedOKCWCBF17} train a multiscale autoregressive model which gradually upsamples images, starting from 4$\times$4 pixels, and makes some independence assumptions in the process. Conditioned on textual captions and spatial keypoints, the model is capable of producing realistic 256$\times$256 images of birds. Although using low-resolution images as representations that abstract away local detail is appealing, this necessarily removes any high-frequency information. Learning these representations instead is more flexible and allows for capturing high-frequency structure.

\citet{menick2018generating} train a variant of PixelCNN dubbed `subscale pixel network' (SPN), which uses a different, hierarchical ordering of the pixels rather than the raster scan order to factorise the joint distribution into conditionals. Their best models consist of two separate SPNs, where one models only the 3 most significant bits of the intensity values at a lower resolution, and another conditionally fills in the remaining information. Trained on the ImageNet dataset, this model is able to generate visually compelling unconditional samples at 128$\times$128 resolution.

The idea of sequential, separate training of levels in a hierarchy dates back to the early days of deep learning~\cite{NIPS2006_3048,Hinton:2006:FLA:1161603.1161605,vincent2010stacked}. \citet{NIPS2018_8023} train a hierarchical autoregressive model of musical audio signals by stacking autoregressive discrete autoencoders. However, the autoencoders are trained end-to-end, which makes them prone to the issues described in Section~\ref{sec:autoregressive-autoencoders}.  Training the second level autoencoder is cumbersome, requiring expensive population-based training~\cite{DBLP:journals/corr/abs-1711-09846} or alternative quantisation strategies to succeed. Applied to images, it leads to colour information being ignored.

\citet{zforcing} use an auxiliary reconstruction loss to avoid posterior collapse in recurrent models. Masked self-prediction is closely related to representation learning methods such as context prediction~\cite{doersch2015unsupervised} and context encoders~\cite{pathak2016context}, which also rely on predicting pixels from other nearby pixels. Contrastive predictive coding~\cite{oord2018representation} on the other hand relies on prediction in the feature domain to extract structure that varies predictably across longer ranges. Although the motivation behind approaches such as these is usually to extract high-level, semantically meaningful features, our goal is different: we want to remove some of the local detail to make the task of modelling large-scale structure easier. We achieve this by predicting only the middle pixel of the masked-out region, which is the main difference compared to previous work. Our representations also need to balance abstraction with reconstruction, so they need to retain enough information from the input.

Context prediction is also a popular representation learning approach in natural language processing, with well-known examples such as word2vec~\cite{NIPS2013_5021}, ELMo~\cite{N18-1202} and BERT~\cite{DBLP:journals/corr/abs-1810-04805}. Other related work includes PixelNet~\cite{PixelNet}, which uses loss functions defined on subsets of image pixels (much like MSP) to tackle dense prediction tasks such as edge detection and semantic segmentation.

In concurrent work, \citet{vqvae2} present a hierarchical generative model of images based on a multi-scale VQ-VAE model combined with one or more autoregressive priors. All levels of latent representations depend directly on the pixels and differ only in their spatial resolution. The pixel-level decoder of this model is feed-forward rather than autoregressive, which enables faster sampling but also results in some degree of blurriness. While this can be an acceptable trade-off for image modelling, accurately capturing high-frequency structure may be important for other data modalities.

\section{Evaluation}
\label{sec:evaluation}
Likelihood-based models can typically be evaluated simply by measuring the likelihood in the pixel domain on a held-out set of images.  With the hierarchical approach, however, different parts of the model are trained using likelihoods measured in different feature spaces, so they are not directly comparable. For a given image, we can measure the likelihood in the feature space modelled by the prior, as well as conditional likelihoods in the domains modelled by each autoregressive decoder, and use these to calculate a joint likelihood across all levels of the model. We can use this as a lower bound for the marginal likelihood (see appendix), but in practice it is dominated by the conditional likelihood of the pixel-level decoder, so it is not particularly informative. Although we report this bound for some models, we stress that likelihoods measured in the pixel domain are not suitable for measuring whether a model captures large-scale structure~\cite{Theis2016a} -- indeed, this is the primary motivation behind our hierarchical approach.

Implicit generative models such as GANs (for which computing likelihoods is intractable) are commonly evaluated using metrics that incorporate pre-trained discriminative models, such as the Inception Score (IS)~\cite{salimans2016improved} and the Fr\'echet Inception Distance (FID)~\cite{heusel2017gans}. Both require a large number of samples, so they are expensive to compute. Nevertheless, sampling from our models is fast enough for this to be tractable (see appendix). These metrics are not without problems however~\cite{barratt2018note,binkowski2018demystifying}, and it is unclear if they correlate well with human perception when used to evaluate non-adversarial generative models.

To complement these computational metrics, we also use human evaluation. We conduct two types of experiments to assess the realism of the generated images, and compare our results to state of the art models: we ask participants to rate individual images for realism on a scale of 1 to 5 and also to compare pairs of images and select the one which looks the most realistic. Note that these experiments are not suitable for assessing diversity, which is much harder to measure.

To allow for further inspection and evaluation, we have made samples for all classes that were used for evaluation available at \url{https://bit.ly/2FJkvhJ}. When generating samples, we can change the temperature of the multinomial distribution which we sequentially sample from for each channel and spatial position. We find that slightly reducing the temperature from the default of $1.0$ to $0.97$ or $0.98$ more consistently yields high quality samples.

\section{Experiments and results}
\label{sec:experiments-and-results}

All experiments were performed on the ImageNet~\cite{deng2009imagenet} and Downsampled ImageNet~\cite{imnet64x64} datasets, with full bit-depth RGB images (8 bits per channel)\footnote{We use the splits (training, validation and testing) as defined by the dataset creators.}. For experiments on resolutions of 128$\times$128 and 256$\times$256, we rescale the shortest side of the image to the desired size and then randomly crop a square image during training (to preserve the aspect ratio). We also use random flipping and brightness, saturation and contrast changes to augment the training data~\cite{43022}. For 64$\times$64 experiments, we use the images made available by \citet{pmlr-v48-oord16}, without augmentation.

\subsection{Auxiliary decoder design}
\label{sec:auxiliary-decoder-design}
As the auxiliary decoders are responsible for shaping the representations learnt by the encoders, their architecture can be varied to influence the information content. We use residual networks for both the encoders and auxiliary decoders~\cite{He2016DeepRL} and vary the number of layers. Each additional layer extends the receptive field of the decoder, which implies that a larger neighbourhood in the code space can affect any given spatial position in the output space. As a result, the information about each pixel is spread out across a larger neighbourhood in the code space, which allows for more efficient use of the discrete bottleneck.

To measure the effect this has on the compressibility of the codes, we first train some autoencoder models on 64$\times$64 colour images using a discrete bottleneck with a single 8-bit channel (256 codes) and downsampling to  32$\times$32 using a strided convolutional layer. The codes are upsampled in the decoder using a subpixel convolutional layer~\cite{DBLP:conf/cvpr/ShiCHTABRW16}. We then train prior models and measure the validation negative log-likelihood (NLL) they achieve at the end of training. The priors are modestly sized PixelCNN models with 20 layers and 128 units per layer.

\begin{wrapfigure}{R}{0.50\textwidth}
  \centering
  \includegraphics[width=1.0\linewidth,trim={0cm 0cm 0cm 0cm},clip]{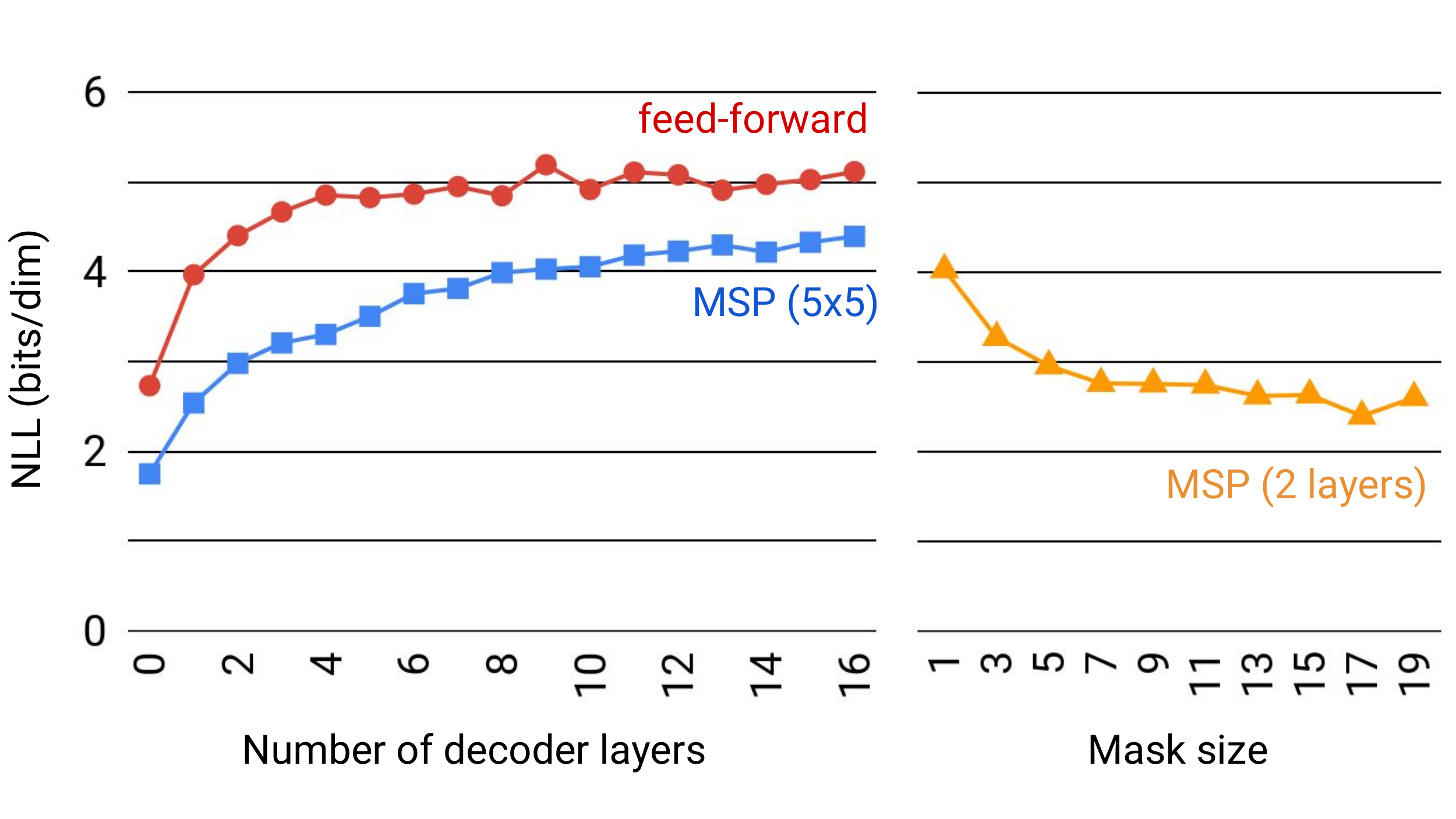}
  \vskip -0.15in
  \caption{Code predictability for different encoders, as measured by the validation NLL of a small PixelCNN prior (see text for details). Left: increasing the number of auxiliary decoder layers makes the codes harder to predict. NLLs for codes from feed-forward decoders (red circles) flatten out more quickly than those from MSP decoders (mask size 5$\times$5, blue squares). Right: increasing the mask size for MSP decoders makes the resulting codes easier to predict (orange triangles).}
  \label{fig:predictability}
  \vskip -0.30in
\end{wrapfigure} 
The results for both feed-forward and MSP decoders (mask size 5$\times$5) are shown in Figure~\ref{fig:predictability} (left). It is clear that the codes become less predictable as the receptive field of the auxiliary decoder increases. As expected, the MSP codes are also more predictable than the feed-forward codes. The predictability of the feed-forward codes seems to flatten out after about 8 decoder layers, while that of the MSP codes decreases more gradually.

For MSP decoders, we repeat this experiment fixing the number of layers to 2 and varying the mask size instead (Figure~\ref{fig:predictability}, right). As expected, increasing the mask size reduces the information content of the codes and makes them more predictable. In the appendix, we also discuss the effect of the auxiliary decoder design on reconstruction quality.

\subsection{Codes abstract away local detail}
\label{sec:abstraction-local-detail}

In this subsection and the next, we evaluate the building blocks that we will use to construct hierarchical models of 128$\times$128 and 256$\times$256 RGB images. To verify that the codes learnt using auxiliary decoders abstract away local detail, we show the variability of sampled reconstructions. Note that the codes for a given input are deterministic, so all stochasticity comes from the autoregressive decoder. We compress all images into single-channel 8-bit codes (256 bins) at 32$\times$32 resolution. We could obtain higher-fidelity reconstructions by increasing the code capacity (by adding more channels, increasing their bit-depth or increasing the resolution), but we choose to use single-channel codes with a modest capacity, so that we can train powerful priors that do not require code channel masking to ensure causality. 

\begin{figure}
    \centering
    \includegraphics[width=0.8\linewidth]{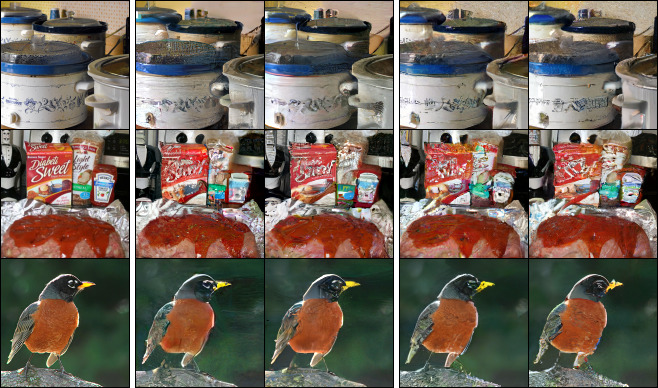}
    \vskip -0.10in
    \caption{Autoregressive autoencoder reconstructions of 128$\times$128 images. Left: original images. Middle: two different sampled reconstructions from models with a feed-forward auxiliary decoder trained with the MSE loss. Right: two different sampled reconstructions from models with an MSP auxiliary decoder with mask size 3$\times$3. The sampling temperature was $0.99$. More examples can be found in the appendix.
    }
    \label{fig:reconstructions-128x128}
    \vskip -0.15in
\end{figure}

For 128$\times$128 images (48$\times$ compression\footnote{From $128\times 128 \times3 \times8$ bits to $32\times 32\times 1\times 8$ bits}), we use a feed-forward auxiliary decoder with 8 layers trained with the MSE loss, and an MSP decoder with 8 layers and a 3$\times$3 mask size. Reconstructions for both are shown in Figure~\ref{fig:reconstructions-128x128}. Note how the MSE decoder is better at preserving local structure (e.g. text on the box), while the MSP decoder is better at preserving the presence of textures (e.g. wallpaper, body of the bird). Both types of reconstructions show variation in colour and texture.

\begin{figure*}
    \centering
    \includegraphics[width=1.0\linewidth]{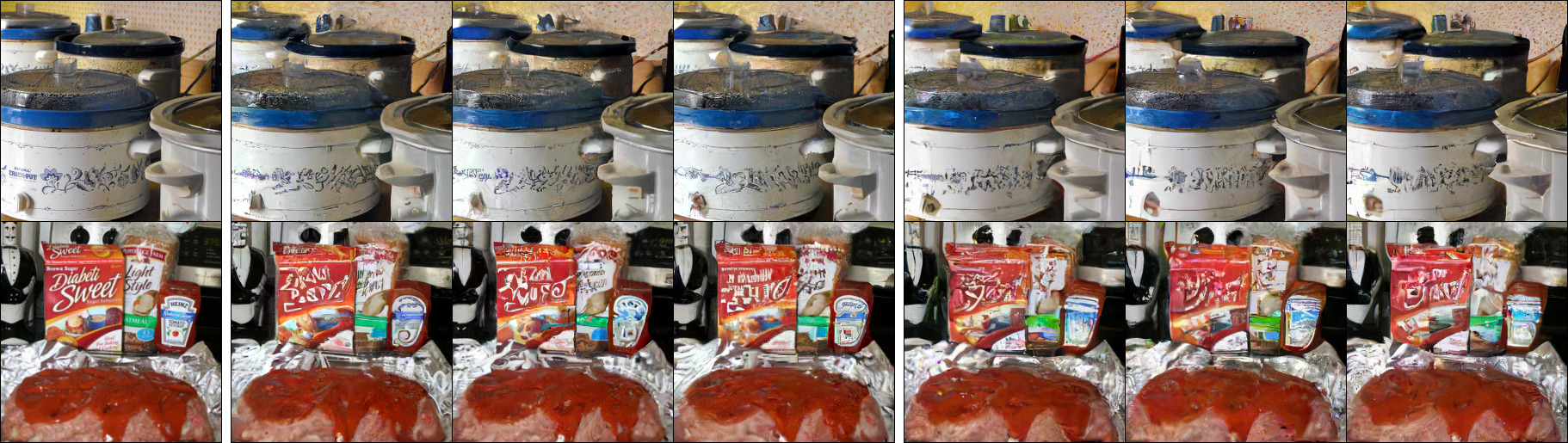}
    \vskip -0.10in
    \caption{Autoregressive autoencoder reconstructions of 256$\times$256 images. Left: original images. Middle: three different sampled two-level reconstructions from models with a feed-forward auxiliary decoder trained with the MSE loss. Right: three different sampled two-level reconstructions from models with an MSP auxiliary decoder with mask sizes 5$\times$5 (level 1) and 3$\times$3 (level 2). The sampling temperature was $0.99$. More examples can be found in the appendix.
    }
    \label{fig:reconstructions-256x256}
    \vskip -0.15in
\end{figure*}

For 256$\times$256 images (192$\times$ compression), we use a stack of two autoencoders, because using a single autoencoder would result in a significant degradation in visual fidelity (see appendix). In Figure~\ref{fig:reconstructions-256x256}, we show reconstructions from a stack trained with feed-forward auxiliary decoders, where the first level model compresses to single-channel 8-bit codes at 128$\times$128 resolution, so the first and second level models compress by factors of 12$\times$ and 16$\times$ respectively (using 1 and 12 auxiliary decoder layers respectively). We also show reconstructions from a stack trained with MSP auxiliary decoders, where the first level compresses to 8-bit codes at 64$\times$64 resolution with 3 channels, so the first and second level models compress by factors of 16$\times$ (with 4 auxiliary decoder layers) and 12$\times$ (with 8 layers) respectively. We refer to the appendix for an exploration of the information captured in the codes as a function of the hyperparameters of the auxiliary decoders.

\subsection{Hierarchical models}
\label{sec:experiments-priors}
We construct hierarchical models using the autoencoders from the previous section, by training class-conditional autoregressive priors on the codes they produce. Once a prior is trained, we can use ancestral sampling to generate images. Most of the model capacity in our hierarchical models should be used for visually salient large-scale structure, so we use powerful prior models, while the decoder models are relatively small.

\begin{figure}
  \centering
  \includegraphics[width=0.8\linewidth]{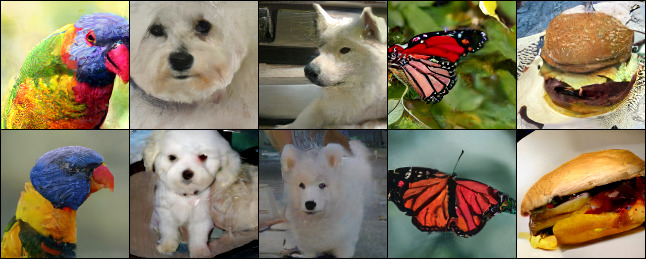}
  \caption{Selected class conditional 128$\times$128 samples from our models with feed-forward auxiliary decoders (top) and MSP decoders (bottom). More are available in the appendix and at \url{https://bit.ly/2FJkvhJ}.}
  \label{fig:128_samples}
  \vskip -0.15in
\end{figure}

\paragraph{128$\times$128 images.} We report results and show samples for two priors: one trained on the feed-forward codes from the previous section, and one for the MSP codes. The prior models are large gated PixelCNNs augmented with masked self-attention layers~\cite{NIPS2017_7181}, which are inserted after every few convolutional layers as in PixelSNAIL~\cite{DBLP:conf/icml/ChenMRA18}. Selected samples are shown in Figure~\ref{fig:128_samples}. Samples for all classes are available at \url{https://bit.ly/2FJkvhJ}. IS and FID are reported in Table~\ref{tab:fid-is-nll-bound}, as well as the joint NLL over the pixels and codes. We do not directly compare with results from previous papers as we cannot compute exact likelihoods, and differences in preprocessing of the images can significantly affect these measurements. 
The IS and FID are much worse than those reported for recent adversarial models~\cite{brock2018large}, but they are in the same ballpark as those reported for PixelCNN on 32$\times$32 ImageNet by \citet{DBLP:conf/icml/OstrovskiDM18} (IS 8.33, FID 33.27).

\begin{table}
\caption{IS and FID for autoregressive priors trained on 32x32 codes obtained from 128x128 images. We also report the joint NLL as discussed in Section~\ref{sec:evaluation}.}
\label{tab:fid-is-nll-bound}
\begin{center}
\begin{small}
\begin{tabular}{rccc}
\toprule
\sc{Aux. decoder}    & \sc{IS}          & \sc{FID} & \sc{joint NLL} \\
\midrule
Feed-forward         & 18.10~$\pm$~0.96 & 44.95    & 3.343 bits/dim \\  
MSP                  & 17.02~$\pm$~0.79 & 46.05    & 3.409 bits/dim \\

\bottomrule
\end{tabular}
\end{small}
\end{center}
\vskip -0.1in
\end{table}

\paragraph{256$\times$256 images.} We trained two priors, one on feed-forward codes and one on MSP codes. Samples from both are available in the appendix and at \url{https://bit.ly/2FJkvhJ}.

\paragraph{Human evaluation.} We asked human raters to rate 128$\times$128 images from 20 different classes generated by different models on a scale from 1 to 5 in terms of realism (Table~\ref{tab:human-eval-results-ratings}). Note that even real images get a relatively low rating on this scale due to the limited resolution. In line with expectations, our hierarchical models receive lower realism scores than BigGAN and are on par with subscale pixel networks. We also asked human raters to compare samples from certain models side by side and pick the most realistic looking ones (Table~\ref{tab:human-eval-results-comp}). Here, we get similar results: samples from a hierarchical model are preferred over BigGAN samples in 22.89\% of cases, and over real images in just 5.39\% of cases. Details of our experimental setup can be found in the appendix, as well as a few nearest neighbour comparisons of samples with images from the dataset in different feature spaces.

\begin{table}
\caption{Human realism ratings (from 1 to 5) for samples from different models. We report the average rating and standard error.}
\label{tab:human-eval-results-ratings}
\begin{center}
\begin{small}
\begin{tabular}{lc}
\toprule
\sc{Model} & \sc{Average rating} \\
\midrule
real images & $2.941 \pm 0.027$ \\
\midrule
BigGAN (high truncation)~\cite{brock2018large} & $2.113 \pm 0.024$ \\
BigGAN (low truncation)~\cite{brock2018large} & $1.874 \pm 0.022$ \\
Subscale pixel network~\cite{menick2018generating} & $1.452 \pm 0.017$ \\
\midrule
Hierarchical (MSP, ours) & $1.431 \pm 0.016$ \\
Hierarchical (feed-forward, ours) & $1.412 \pm 0.015$ \\
\bottomrule
\end{tabular}
\end{small}
\end{center}
\end{table}

\begin{table}
\caption{Human rater preference in a pairwise comparison experiment between samples from different models.}
\label{tab:human-eval-results-comp}
\begin{center}
\begin{small}
\begin{tabular}{ccccc}
\toprule
\multicolumn{2}{c}{\sc{Model}}          & \multicolumn{3}{c}{\sc{Preference}} \\
\sc{A} & \sc{B}                         & \sc{A} & \sc{B} & \sc{Unknown} \\
\midrule
MSP (ours) & Feed-forward (ours)        & $50.89\%$ & $48.93\%$ & $0.18\%$ \\
\midrule
MSP (ours) & SPN                        & $50.57\%$ & $49.38\%$ & $0.05\%$ \\
MSP (ours) & BigGAN (low truncation)    & $22.89\%$ & $77.01\%$ & $0.10\%$ \\
MSP (ours) & real images                & $5.39\%$ & $94.44\%$ & $0.17\%$ \\
\bottomrule
\end{tabular}
\end{small}
\end{center}
\end{table}

\section{Conclusion}
\label{sec:conclusion}

We have discussed the challenges of training autoregressive autoencoders, and proposed two techniques that address these challenges using auxiliary decoders. The first uses a feed-forward network to reconstruct pixels, while the second relies on predicting missing pixels. This enabled us to build hierarchical autoregressive models of images which are capable of producing high-fidelity class-conditional samples with large-scale coherence at resolutions of 128$\times$128 and 256$\times$256 when trained on ImageNet. This demonstrates that our hierarchical approach can be used to effectively scale up likelihood-based generative models. In future work, we would like to compare both techniques in more challenging settings and further explore their relative strengths and limitations.

\subsubsection*{Acknowledgments}
We would like to thank the following people for their help and input: A\"aron van den Oord, Ali Razavi, Jacob Menick, Marco Cornero, Tamas Berghammer, Andy Brock, Jeff Donahue, Carl Doersch, Jacob Walker, Chloe Hillier, Louise Deason, Scott Reed, Nando de Freitas, Mary Chesus, Jonathan Godwin, Trevor Back and Anish Athalye.

\small
\bibliographystyle{plainnat}
\bibliography{paper}

\begin{thebibliography}{56}
\providecommand{\natexlab}[1]{#1}
\providecommand{\url}[1]{\texttt{#1}}
\expandafter\ifx\csname urlstyle\endcsname\relax
  \providecommand{\doi}[1]{doi: #1}\else
  \providecommand{\doi}{doi: \begingroup \urlstyle{rm}\Url}\fi

\bibitem[Bansal et~al.(2017)Bansal, Chen, Russell, Gupta, and
  Ramanan]{PixelNet}
Aayush Bansal, Xinlei Chen, Bryan Russell, Abhinav Gupta, and Deva Ramanan.
\newblock Pixelnet: Representation of the pixels, by the pixels, and for the
  pixels.
\newblock \emph{arXiv:1702.06506}, 2017.

\bibitem[Barratt and Sharma(2018)]{barratt2018note}
Shane Barratt and Rishi Sharma.
\newblock A note on the inception score.
\newblock \emph{arXiv preprint arXiv:1801.01973}, 2018.

\bibitem[Bengio et~al.(2007)Bengio, Lamblin, Popovici, and
  Larochelle]{NIPS2006_3048}
Yoshua Bengio, Pascal Lamblin, Dan Popovici, and Hugo Larochelle.
\newblock Greedy layer-wise training of deep networks.
\newblock In B.~Sch\"{o}lkopf, J.~C. Platt, and T.~Hoffman, editors,
  \emph{Advances in Neural Information Processing Systems 19}, pages 153--160.
  MIT Press, 2007.
\newblock URL
  \url{http://papers.nips.cc/paper/3048-greedy-layer-wise-training-of-deep-networks.pdf}.

\bibitem[Bengio et~al.(2013)Bengio, L{\'e}onard, and
  Courville]{bengio2013estimating}
Yoshua Bengio, Nicholas L{\'e}onard, and Aaron Courville.
\newblock Estimating or propagating gradients through stochastic neurons for
  conditional computation.
\newblock \emph{arXiv preprint arXiv:1308.3432}, 2013.

\bibitem[Bińkowski et~al.(2018)Bińkowski, Sutherland, Arbel, and
  Gretton]{binkowski2018demystifying}
Mikołaj Bińkowski, Dougal~J. Sutherland, Michael Arbel, and Arthur Gretton.
\newblock Demystifying {MMD} {GAN}s.
\newblock In \emph{International Conference on Learning Representations}, 2018.

\bibitem[Bowman et~al.(2016)Bowman, Vilnis, Vinyals, Dai, J{\'{o}}zefowicz, and
  Bengio]{DBLP:conf/conll/BowmanVVDJB16}
Samuel~R. Bowman, Luke Vilnis, Oriol Vinyals, Andrew~M. Dai, Rafal
  J{\'{o}}zefowicz, and Samy Bengio.
\newblock Generating sentences from a continuous space.
\newblock In \emph{Proceedings of the 20th {SIGNLL} Conference on Computational
  Natural Language Learning, CoNLL 2016, Berlin, Germany, August 11-12, 2016},
  pages 10--21, 2016.

\bibitem[Brock et~al.(2019)Brock, Donahue, and Simonyan]{brock2018large}
Andrew Brock, Jeff Donahue, and Karen Simonyan.
\newblock Large scale {GAN} training for high fidelity natural image synthesis.
\newblock In \emph{International Conference on Learning Representations}, 2019.

\bibitem[Chen et~al.(2016)Chen, Kingma, Salimans, Duan, Dhariwal, Schulman,
  Sutskever, and Abbeel]{DBLP:journals/corr/ChenKSDDSSA16}
Xi~Chen, Diederik~P. Kingma, Tim Salimans, Yan Duan, Prafulla Dhariwal, John
  Schulman, Ilya Sutskever, and Pieter Abbeel.
\newblock Variational lossy autoencoder.
\newblock \emph{CoRR}, abs/1611.02731, 2016.

\bibitem[Chen et~al.(2018)Chen, Mishra, Rohaninejad, and
  Abbeel]{DBLP:conf/icml/ChenMRA18}
Xi~Chen, Nikhil Mishra, Mostafa Rohaninejad, and Pieter Abbeel.
\newblock Pixelsnail: An improved autoregressive generative model.
\newblock In \emph{{ICML}}, volume~80 of \emph{{JMLR} Workshop and Conference
  Proceedings}, pages 863--871. JMLR.org, 2018.

\bibitem[Deng et~al.(2009)Deng, Dong, Socher, Li, Li, and
  Fei-Fei]{deng2009imagenet}
Jia Deng, Wei Dong, Richard Socher, Li-Jia Li, Kai Li, and Li~Fei-Fei.
\newblock Imagenet: A large-scale hierarchical image database.
\newblock In \emph{Computer Vision and Pattern Recognition, 2009. CVPR 2009.
  IEEE Conference on}, pages 248--255. Ieee, 2009.

\bibitem[Devlin et~al.(2018)Devlin, Chang, Lee, and
  Toutanova]{DBLP:journals/corr/abs-1810-04805}
Jacob Devlin, Ming{-}Wei Chang, Kenton Lee, and Kristina Toutanova.
\newblock {BERT:} pre-training of deep bidirectional transformers for language
  understanding.
\newblock \emph{CoRR}, abs/1810.04805, 2018.

\bibitem[Dieleman et~al.(2018)Dieleman, van~den Oord, and
  Simonyan]{NIPS2018_8023}
Sander Dieleman, Aaron van~den Oord, and Karen Simonyan.
\newblock The challenge of realistic music generation: modelling raw audio at
  scale.
\newblock In S.~Bengio, H.~Wallach, H.~Larochelle, K.~Grauman, N.~Cesa-Bianchi,
  and R.~Garnett, editors, \emph{Advances in Neural Information Processing
  Systems 31}, pages 8000--8010. Curran Associates, Inc., 2018.

\bibitem[Dinh et~al.(2014)Dinh, Krueger, and
  Bengio]{DBLP:journals/corr/DinhKB14}
Laurent Dinh, David Krueger, and Yoshua Bengio.
\newblock {NICE:} non-linear independent components estimation.
\newblock \emph{CoRR}, abs/1410.8516, 2014.

\bibitem[Dinh et~al.(2017)Dinh, Sohl-Dickstein, and Bengio]{45819}
Laurent Dinh, Jascha Sohl-Dickstein, and Samy Bengio.
\newblock Density estimation using real nvp.
\newblock 2017.

\bibitem[Doersch et~al.(2015)Doersch, Gupta, and
  Efros]{doersch2015unsupervised}
Carl Doersch, Abhinav Gupta, and Alexei~A Efros.
\newblock Unsupervised visual representation learning by context prediction.
\newblock In \emph{Proceedings of the IEEE International Conference on Computer
  Vision}, pages 1422--1430, 2015.

\bibitem[Engel et~al.(2017)Engel, Resnick, Roberts, Dieleman, Norouzi, Eck, and
  Simonyan]{DBLP:conf/icml/EngelRRDNES17}
Jesse Engel, Cinjon Resnick, Adam Roberts, Sander Dieleman, Mohammad Norouzi,
  Douglas Eck, and Karen Simonyan.
\newblock Neural audio synthesis of musical notes with wavenet autoencoders.
\newblock In \emph{Proceedings of the 34th International Conference on Machine
  Learning, {ICML} 2017, Sydney, NSW, Australia, 6-11 August 2017}, pages
  1068--1077, 2017.

\bibitem[Goodfellow et~al.(2014)Goodfellow, Pouget-Abadie, Mirza, Xu,
  Warde-Farley, Ozair, Courville, and Bengio]{NIPS2014_5423}
Ian Goodfellow, Jean Pouget-Abadie, Mehdi Mirza, Bing Xu, David Warde-Farley,
  Sherjil Ozair, Aaron Courville, and Yoshua Bengio.
\newblock Generative adversarial nets.
\newblock In Z.~Ghahramani, M.~Welling, C.~Cortes, N.~D. Lawrence, and K.~Q.
  Weinberger, editors, \emph{Advances in Neural Information Processing Systems
  27}, pages 2672--2680. Curran Associates, Inc., 2014.

\bibitem[Goyal et~al.(2017)Goyal, Sordoni, Côté, Ke, and Bengio]{zforcing}
Anirudh Goyal, Alessandro Sordoni, Marc-Alexandre Côté, Nan~Rosemary Ke, and
  Yoshua Bengio.
\newblock Z-forcing: Training stochastic recurrent networks.
\newblock pages 6697--6707, November 2017.
\newblock URL
  \url{https://www.microsoft.com/en-us/research/publication/z-forcing-training-stochastic-recurrent-networks/}.

\bibitem[Gulrajani et~al.(2016)Gulrajani, Kumar, Ahmed, Ali~Taiga, Visin,
  Vazquez, and Courville]{gulrajani+al-2016-pixelvae}
Ishaan Gulrajani, Kundan Kumar, Faruk Ahmed, Adrien Ali~Taiga, Francesco Visin,
  David Vazquez, and Aaron Courville.
\newblock {PixelVAE}: A latent variable model for natural images.
\newblock \emph{arXiv e-prints}, abs/1611.05013, November 2016.

\bibitem[He et~al.(2016{\natexlab{a}})He, Zhang, Ren, and Sun]{He2016DeepRL}
Kaiming He, Xiangyu Zhang, Shaoqing Ren, and Jian Sun.
\newblock Deep residual learning for image recognition.
\newblock \emph{2016 IEEE Conference on Computer Vision and Pattern Recognition
  (CVPR)}, pages 770--778, 2016{\natexlab{a}}.

\bibitem[He et~al.(2016{\natexlab{b}})He, Zhang, Ren, and
  Sun]{He2016IdentityMI}
Kaiming He, Xiangyu Zhang, Shaoqing Ren, and Jian Sun.
\newblock Identity mappings in deep residual networks.
\newblock In \emph{ECCV}, 2016{\natexlab{b}}.

\bibitem[Heusel et~al.(2017)Heusel, Ramsauer, Unterthiner, Nessler, and
  Hochreiter]{heusel2017gans}
Martin Heusel, Hubert Ramsauer, Thomas Unterthiner, Bernhard Nessler, and Sepp
  Hochreiter.
\newblock Gans trained by a two time-scale update rule converge to a local nash
  equilibrium.
\newblock In \emph{Advances in Neural Information Processing Systems}, pages
  6626--6637, 2017.

\bibitem[Hinton et~al.(2015)Hinton, Vinyals, and Dean]{44873}
Geoffrey Hinton, Oriol Vinyals, and Jeffrey Dean.
\newblock Distilling the knowledge in a neural network.
\newblock In \emph{NIPS Deep Learning and Representation Learning Workshop},
  2015.

\bibitem[Hinton et~al.(2006)Hinton, Osindero, and
  Teh]{Hinton:2006:FLA:1161603.1161605}
Geoffrey~E. Hinton, Simon Osindero, and Yee-Whye Teh.
\newblock A fast learning algorithm for deep belief nets.
\newblock \emph{Neural Comput.}, 18\penalty0 (7):\penalty0 1527--1554, July
  2006.
\newblock ISSN 0899-7667.
\newblock \doi{10.1162/neco.2006.18.7.1527}.
\newblock URL \url{http://dx.doi.org/10.1162/neco.2006.18.7.1527}.

\bibitem[Jaderberg et~al.(2017)Jaderberg, Dalibard, Osindero, Czarnecki,
  Donahue, Razavi, Vinyals, Green, Dunning, Simonyan, Fernando, and
  Kavukcuoglu]{DBLP:journals/corr/abs-1711-09846}
Max Jaderberg, Valentin Dalibard, Simon Osindero, Wojciech~M. Czarnecki, Jeff
  Donahue, Ali Razavi, Oriol Vinyals, Tim Green, Iain Dunning, Karen Simonyan,
  Chrisantha Fernando, and Koray Kavukcuoglu.
\newblock Population based training of neural networks.
\newblock \emph{CoRR}, abs/1711.09846, 2017.

\bibitem[Karras et~al.(2018{\natexlab{a}})Karras, Aila, Laine, and
  Lehtinen]{karras2018progressive}
Tero Karras, Timo Aila, Samuli Laine, and Jaakko Lehtinen.
\newblock Progressive growing of {GAN}s for improved quality, stability, and
  variation.
\newblock In \emph{International Conference on Learning Representations},
  2018{\natexlab{a}}.

\bibitem[Karras et~al.(2018{\natexlab{b}})Karras, Laine, and Aila]{nvidiastyle}
Tero Karras, Samuli Laine, and Timo Aila.
\newblock A style-based generator architecture for generative adversarial
  networks.
\newblock \emph{CoRR}, abs/1812.04948, 2018{\natexlab{b}}.

\bibitem[Kingma and Welling(2013)]{KingmaW13}
Diederik~P. Kingma and Max Welling.
\newblock Auto-encoding variational bayes.
\newblock \emph{CoRR}, abs/1312.6114, 2013.

\bibitem[Kingma and Dhariwal(2018)]{NIPS2018_8224}
Durk~P Kingma and Prafulla Dhariwal.
\newblock Glow: Generative flow with invertible 1x1 convolutions.
\newblock In S.~Bengio, H.~Wallach, H.~Larochelle, K.~Grauman, N.~Cesa-Bianchi,
  and R.~Garnett, editors, \emph{Advances in Neural Information Processing
  Systems 31}, pages 10236--10245. Curran Associates, Inc., 2018.

\bibitem[Menick and Kalchbrenner(2019)]{menick2018generating}
Jacob Menick and Nal Kalchbrenner.
\newblock Generating high fidelity images with subscale pixel networks and
  multidimensional upscaling.
\newblock In \emph{International Conference on Learning Representations}, 2019.

\bibitem[Mikolov et~al.(2013)Mikolov, Sutskever, Chen, Corrado, and
  Dean]{NIPS2013_5021}
Tomas Mikolov, Ilya Sutskever, Kai Chen, Greg~S Corrado, and Jeff Dean.
\newblock Distributed representations of words and phrases and their
  compositionality.
\newblock In C.~J.~C. Burges, L.~Bottou, M.~Welling, Z.~Ghahramani, and K.~Q.
  Weinberger, editors, \emph{Advances in Neural Information Processing Systems
  26}, pages 3111--3119. Curran Associates, Inc., 2013.

\bibitem[Oord et~al.(2018)Oord, Li, and Vinyals]{oord2018representation}
Aaron van~den Oord, Yazhe Li, and Oriol Vinyals.
\newblock Representation learning with contrastive predictive coding.
\newblock \emph{arXiv preprint arXiv:1807.03748}, 2018.

\bibitem[Ostrovski et~al.(2018)Ostrovski, Dabney, and
  Munos]{DBLP:conf/icml/OstrovskiDM18}
Georg Ostrovski, Will Dabney, and R{\'{e}}mi Munos.
\newblock Autoregressive quantile networks for generative modeling.
\newblock In \emph{{ICML}}, volume~80 of \emph{{JMLR} Workshop and Conference
  Proceedings}, pages 3933--3942. JMLR.org, 2018.

\bibitem[Paine et~al.(2016)Paine, Khorrami, Chang, Zhang, Ramachandran,
  Hasegawa{-}Johnson, and Huang]{DBLP:journals/corr/PaineKCZRHH16}
Tom~Le Paine, Pooya Khorrami, Shiyu Chang, Yang Zhang, Prajit Ramachandran,
  Mark~A. Hasegawa{-}Johnson, and Thomas~S. Huang.
\newblock Fast wavenet generation algorithm.
\newblock \emph{CoRR}, abs/1611.09482, 2016.
\newblock URL \url{http://arxiv.org/abs/1611.09482}.

\bibitem[Pathak et~al.(2016)Pathak, Krahenbuhl, Donahue, Darrell, and
  Efros]{pathak2016context}
Deepak Pathak, Philipp Krahenbuhl, Jeff Donahue, Trevor Darrell, and Alexei~A
  Efros.
\newblock Context encoders: Feature learning by inpainting.
\newblock In \emph{Proceedings of the IEEE Conference on Computer Vision and
  Pattern Recognition}, pages 2536--2544, 2016.

\bibitem[Peters et~al.(2018)Peters, Neumann, Iyyer, Gardner, Clark, Lee, and
  Zettlemoyer]{N18-1202}
Matthew Peters, Mark Neumann, Mohit Iyyer, Matt Gardner, Christopher Clark,
  Kenton Lee, and Luke Zettlemoyer.
\newblock Deep contextualized word representations.
\newblock In \emph{Proceedings of the 2018 Conference of the North American
  Chapter of the Association for Computational Linguistics: Human Language
  Technologies, Volume 1 (Long Papers)}, pages 2227--2237. Association for
  Computational Linguistics, 2018.
\newblock \doi{10.18653/v1/N18-1202}.

\bibitem[Polyak and Juditsky(1992)]{Polyak:1992:ASA:131092.131098}
B.~T. Polyak and A.~B. Juditsky.
\newblock Acceleration of stochastic approximation by averaging.
\newblock \emph{SIAM J. Control Optim.}, 30\penalty0 (4):\penalty0 838--855,
  July 1992.
\newblock ISSN 0363-0129.
\newblock \doi{10.1137/0330046}.
\newblock URL \url{http://dx.doi.org/10.1137/0330046}.

\bibitem[Radford et~al.(2015)Radford, Metz, and
  Chintala]{radford2015unsupervised}
Alec Radford, Luke Metz, and Soumith Chintala.
\newblock Unsupervised representation learning with deep convolutional
  generative adversarial networks.
\newblock \emph{arXiv preprint arXiv:1511.06434}, 2015.

\bibitem[Razavi et~al.(2019{\natexlab{a}})Razavi, Oord, Poole, and
  Vinyals]{razavi2019preventing}
Ali Razavi, A{\"a}ron van~den Oord, Ben Poole, and Oriol Vinyals.
\newblock Preventing posterior collapse with delta-vaes.
\newblock \emph{arXiv preprint arXiv:1901.03416}, 2019{\natexlab{a}}.

\bibitem[Razavi et~al.(2019{\natexlab{b}})Razavi, van~den Oord, and
  Vinyals]{vqvae2}
Ali Razavi, Aaron van~den Oord, and Oriol Vinyals.
\newblock Generating diverse high resolution images with vq-vae.
\newblock \emph{ICLR 2019 Workshop DeepGenStruct}, 2019{\natexlab{b}}.
\newblock URL \url{https://openreview.net/forum?id=ryeBN88Ku4}.

\bibitem[Reed et~al.(2017)Reed, van~den Oord, Kalchbrenner, Colmenarejo, Wang,
  Chen, Belov, and de~Freitas]{DBLP:conf/icml/ReedOKCWCBF17}
Scott~E. Reed, A{\"{a}}ron van~den Oord, Nal Kalchbrenner, Sergio~Gomez
  Colmenarejo, Ziyu Wang, Yutian Chen, Dan Belov, and Nando de~Freitas.
\newblock Parallel multiscale autoregressive density estimation.
\newblock In \emph{{ICML}}, volume~70 of \emph{Proceedings of Machine Learning
  Research}, pages 2912--2921. {PMLR}, 2017.

\bibitem[Rezende et~al.(2014)Rezende, Mohamed, and
  Wierstra]{pmlr-v32-rezende14}
Danilo~Jimenez Rezende, Shakir Mohamed, and Daan Wierstra.
\newblock Stochastic backpropagation and approximate inference in deep
  generative models.
\newblock In Eric~P. Xing and Tony Jebara, editors, \emph{Proceedings of the
  31st International Conference on Machine Learning}, volume~32 of
  \emph{Proceedings of Machine Learning Research}, pages 1278--1286, Bejing,
  China, 22--24 Jun 2014. PMLR.

\bibitem[Salimans et~al.(2016)Salimans, Goodfellow, Zaremba, Cheung, Radford,
  and Chen]{salimans2016improved}
Tim Salimans, Ian Goodfellow, Wojciech Zaremba, Vicki Cheung, Alec Radford, and
  Xi~Chen.
\newblock Improved techniques for training gans.
\newblock In \emph{Advances in Neural Information Processing Systems}, pages
  2234--2242, 2016.

\bibitem[Shi et~al.(2016)Shi, Caballero, Huszar, Totz, Aitken, Bishop,
  Rueckert, and Wang]{DBLP:conf/cvpr/ShiCHTABRW16}
Wenzhe Shi, Jose Caballero, Ferenc Huszar, Johannes Totz, Andrew~P. Aitken, Rob
  Bishop, Daniel Rueckert, and Zehan Wang.
\newblock Real-time single image and video super-resolution using an efficient
  sub-pixel convolutional neural network.
\newblock In \emph{{CVPR}}, pages 1874--1883. {IEEE} Computer Society, 2016.

\bibitem[Simonyan and Zisserman(2015)]{Simonyan15}
K.~Simonyan and A.~Zisserman.
\newblock Very deep convolutional networks for large-scale image recognition.
\newblock In \emph{International Conference on Learning Representations}, 2015.

\bibitem[Szegedy et~al.(2015)Szegedy, Liu, Jia, Sermanet, Reed, Anguelov,
  Erhan, Vanhoucke, and Rabinovich]{43022}
Christian Szegedy, Wei Liu, Yangqing Jia, Pierre Sermanet, Scott Reed, Dragomir
  Anguelov, Dumitru Erhan, Vincent Vanhoucke, and Andrew Rabinovich.
\newblock Going deeper with convolutions.
\newblock In \emph{Computer Vision and Pattern Recognition (CVPR)}, 2015.

\bibitem[Theis et~al.(2016)Theis, van~den Oord, and Bethge]{Theis2016a}
L.~Theis, A.~van~den Oord, and M.~Bethge.
\newblock A note on the evaluation of generative models.
\newblock In \emph{International Conference on Learning Representations}, Apr
  2016.

\bibitem[Theis and Bethge(2015)]{NIPS2015_5637}
Lucas Theis and Matthias Bethge.
\newblock Generative image modeling using spatial lstms.
\newblock In C.~Cortes, N.~D. Lawrence, D.~D. Lee, M.~Sugiyama, and R.~Garnett,
  editors, \emph{Advances in Neural Information Processing Systems 28}, pages
  1927--1935. Curran Associates, Inc., 2015.

\bibitem[van~den Oord et~al.()van~den Oord, Kalchbrenner, and
  Kavukcuoglu]{imnet64x64}
Aaron van~den Oord, Nal Kalchbrenner, and Koray Kavukcuoglu.
\newblock Downsampled imagenet 64x64.
\newblock URL \url{http://image-net.org/small/download.php}.

\bibitem[van~den Oord et~al.(2016{\natexlab{a}})van~den Oord, Kalchbrenner,
  Espeholt, kavukcuoglu, Vinyals, and Graves]{NIPS2016_6527}
Aaron van~den Oord, Nal Kalchbrenner, Lasse Espeholt, koray kavukcuoglu, Oriol
  Vinyals, and Alex Graves.
\newblock Conditional image generation with pixelcnn decoders.
\newblock In D.~D. Lee, M.~Sugiyama, U.~V. Luxburg, I.~Guyon, and R.~Garnett,
  editors, \emph{Advances in Neural Information Processing Systems 29}, pages
  4790--4798. Curran Associates, Inc., 2016{\natexlab{a}}.

\bibitem[van~den Oord et~al.(2016{\natexlab{b}})van~den Oord, Kalchbrenner, and
  Kavukcuoglu]{pmlr-v48-oord16}
Aaron van~den Oord, Nal Kalchbrenner, and Koray Kavukcuoglu.
\newblock Pixel recurrent neural networks.
\newblock In Maria~Florina Balcan and Kilian~Q. Weinberger, editors,
  \emph{Proceedings of The 33rd International Conference on Machine Learning},
  volume~48 of \emph{Proceedings of Machine Learning Research}, pages
  1747--1756, New York, New York, USA, 20--22 Jun 2016{\natexlab{b}}. PMLR.

\bibitem[van~den Oord et~al.(2017)van~den Oord, Vinyals, and
  kavukcuoglu]{NIPS2017_7210}
Aaron van~den Oord, Oriol Vinyals, and koray kavukcuoglu.
\newblock Neural discrete representation learning.
\newblock In I.~Guyon, U.~V. Luxburg, S.~Bengio, H.~Wallach, R.~Fergus,
  S.~Vishwanathan, and R.~Garnett, editors, \emph{Advances in Neural
  Information Processing Systems 30}, pages 6306--6315. Curran Associates,
  Inc., 2017.

\bibitem[Vaswani et~al.(2017)Vaswani, Shazeer, Parmar, Uszkoreit, Jones, Gomez,
  Kaiser, and Polosukhin]{NIPS2017_7181}
Ashish Vaswani, Noam Shazeer, Niki Parmar, Jakob Uszkoreit, Llion Jones,
  Aidan~N Gomez, \L~ukasz Kaiser, and Illia Polosukhin.
\newblock Attention is all you need.
\newblock In I.~Guyon, U.~V. Luxburg, S.~Bengio, H.~Wallach, R.~Fergus,
  S.~Vishwanathan, and R.~Garnett, editors, \emph{Advances in Neural
  Information Processing Systems 30}, pages 5998--6008. Curran Associates,
  Inc., 2017.

\bibitem[Vincent et~al.(2010)Vincent, Larochelle, Lajoie, Bengio, and
  Manzagol]{vincent2010stacked}
Pascal Vincent, Hugo Larochelle, Isabelle Lajoie, Yoshua Bengio, and
  Pierre-Antoine Manzagol.
\newblock Stacked denoising autoencoders: Learning useful representations in a
  deep network with a local denoising criterion.
\newblock \emph{Journal of machine learning research}, 11\penalty0
  (Dec):\penalty0 3371--3408, 2010.

\bibitem[Williams and Zipser(1989)]{williams:recurrent}
Ronald~J. Williams and David Zipser.
\newblock A learning algorithm for continually running fully recurrent neural
  networks.
\newblock \emph{Neural Computation}, 1:\penalty0 270--280, 1989.

\bibitem[Zhou et~al.(2019)Zhou, Gordon, Krishna, Narcomey, Morina, and
  Bernstein]{DBLP:journals/corr/abs-1904-01121}
Sharon Zhou, Mitchell Gordon, Ranjay Krishna, Austin Narcomey, Durim Morina,
  and Michael~S. Bernstein.
\newblock {HYPE:} human eye perceptual evaluation of generative models.
\newblock \emph{CoRR}, abs/1904.01121, 2019.
\newblock URL \url{http://arxiv.org/abs/1904.01121}.

\end{thebibliography}

\newpage
\appendix

\section{Architectural details}
In this section, we describe the architecture of the different components of our models, and provide hyperparameters for all experiments. All hierarchical models consist of one or two autoencoder models and a prior, which are trained separately and in sequence. We use Polyak averaging~\cite{Polyak:1992:ASA:131092.131098} for all models with a decay constant of 0.9999.

Each autoencoder consists of a number of subnetworks: an encoder, an autoregressive decoder and an auxiliary decoder. A quantisation bottleneck is inserted between the encoder and both decoders. In the case of MSP training, there is also an additional teacher network. The autoregressive decoder in turn consists of a modulator and a local model. The local model is always a gated PixelCNN~\cite{NIPS2016_6527}. The modulator is a residual net and is responsible for mapping the code input to a set of biases for each layer in the local model. The encoder, auxiliary decoder, and teacher networks are all residual nets as well. For all residual networks, we use the ResNet v2 `full pre-activation' formulation~\cite{He2016IdentityMI} (without batch normalisation), where each residual block consists of a ReLU nonlinearity, a 3$\times$3 convolution, another ReLU nonlinearity and a 1$\times$1 convolution, in that order. Note that we chose not to condition any of the autoencoder components on class labels in our experiments (only the priors are class-conditional).

We train all models on 64$\times$64 crops, unless the input representations already have a resolution of 64$\times$64 or smaller. When training MSP models, we use a different number of masks per image, depending on the mask size. For mask sizes 1$\times$1 and 3$\times$3 we use 30 masks. For mask sizes 5$\times$5 and 7$\times$7 we use 10 masks. For sizes 9, 11, 13 and 15 we use 3 masks, and for 17 and 19 we us a single mask per image. In preliminary experiments, these settings enabled us to get the best self-prediction likelihoods. Note that we present the images and codes to the MSP teacher and encoder in a one-hot representation, so that masked pixels ($[0, 0, \ldots, 0]$) can be distinguished from black pixels ($[1, 0, \ldots, 0]$).

We first describe the architecture details for the autoregressive autoencoders with feed-forward and masked self-prediction auxiliary decoders for the different resolutions. The priors on the resulting codes are described jointly for all models in Section~\ref{sec:prior-details}. 

\subsection{2-level models for 128$\times$128 images}
\subsubsection{With feed-forward decoder}
\label{two-level-ff-128}
The encoder, auxiliary decoder, and modulator are all residual networks with 512 hiddens and a residual bottleneck of 128 hiddens. The encoder and modulator both have 16 layers whereas the auxiliary decoder has only 2 layers. In the encoder the features are downscaled at the end using a strided convolution with a stride of 2. In the auxiliary decoder and the modulator the upsampling (by a factor of 2) is done in the beginning using subpixel convolutional layer~\cite{DBLP:conf/cvpr/ShiCHTABRW16}. The local model, a gated PixelCNN, has 16 layers with 128 units. The VQ bottleneck has 1 channel with 9 bits\footnote{We use 9 bits instead of the 8 bits used in other experiments because there was a significant increase in reconstruction quality when using 9 bits instead of 8.} (512 bins). The model was trained with the Adam optimizer for 300000 iterations.

\subsubsection{With MSP decoder}
\label{two-level-msp-128}
The teacher, encoder, decoder and modulator are all residual networks with 128 hiddens. The encoder, teacher and modulator have 16 layers whereas the auxiliary decoder has 8 layers. We use a mask size of 3$\times$3. The local model has 20 layers and the VQ bottleneck has 1 channel with 8 bits. The model was trained with the Adam optimizer for 200000 iterations. Other than that, the setup matches the one used for the feed-forward decoder.

\subsection{3-level models for 256$\times$256 images}
\subsubsection{With feed-forward decoder}
For the first level model, which maps 256$\times$256 RGB images to 128$\times$128 single channel codes with 8 bits, we can make the components relatively small. We use the same model as described in Section \ref{two-level-ff-128} except for the following: we use only 4 encoder layers, 1 layer in the auxiliary decoder and 8 autoregressive layers.

The second level model, which maps the 128$\times$128 codes to single channel 32$\times$32 codes with 8 bits, uses 16 encoder and 16 modulator layers with 1024 hiddens and residual bottlenecks of 256 hiddens. The auxiliary decoder has 12 layers and the autoregressive decoder is also only 8 layers but now has 384 hiddens per layer.

\subsubsection{With MSP decoder}
For the first level model, which maps 256$\times$256 RGB images to 64$\times$64 3-channel codes with 8 bits, we use the same model as described in Section~\ref{two-level-msp-128}, except that the auxiliary decoder has 4 layers and the mask size is 5$\times$5.

The second level model, which maps the 64$\times$64 codes to single channel 32$\times$32 codes with 8 bits is also the same, but has an auxiliary decoder with 8 layers and the mask size is 3$\times$3.

\subsection{Prior details}
\label{sec:prior-details}
Our autoregressive priors are very similar to those described by \citet{razavi2019preventing}, which are in turn closely related to PixelSNAIL \cite{DBLP:conf/icml/ChenMRA18}.
We list their details in Table \ref{tab:prior-arch-details}.

\begin{table}
\caption{Architecture details for the priors on the different codes: FF denotes the feed-forward decoder model and MSP denotes the masked self-prediction model. $l$ is the number of layers, $h$ the number of hiddens for each layer, $r$ is residual filter size, $timing$ denotes if the timing signal was added or concatenated with the input, $a$ is number of attention layers, $ah$ is the number of attention heads, $do$ is the probability of dropout, $bs$ is the batch size and $iters$ is the number of iterations the model has been trained for.}
\label{tab:prior-arch-details}
\begin{center}
\begin{small}
\begin{tabular}{rcccc}
\toprule
& \sc{128$\times$128}      & \sc{128$\times$128} & \sc{256$\times$256}  & \sc{256$\times$256} \\
& \sc{FF}         & \sc{MSP} & \sc{FF}  & \sc{MSP} \\
\midrule
$l$         & 20 & 20    & 20  & 20 \\ 
$h$      & 640 & 640    & 640  & 640 \\
$r$   & 2048 & 2048    & 2048  & 2048 \\
$timing$   & add &  add   &  concat & add \\
$a$   & 6 & 5 & 5  & 5 \\
$ah$   & 10 & 10 & 15 & 10  \\
$do$   & 0.2 & 0.1 &  0.0 & 0.1  \\
$bs$   & 2048 & 2048 & 2048 & 2048 \\ 
$iters$   & 490200 & 267000 &  422700 & 185000 \\

\bottomrule
\end{tabular}
\end{small}
\end{center}
\vskip -0.1in
\end{table}

\section{Training}
All models were trained on Google TPU v3 Pods. The encoders and decoders were trained on pods with 32 cores, whereas the priors have been trained on pods with 512 cores.

The procedure for training hierarchical models is outlined in Algorithm~\ref{algo:training}. The procedures for training encoders with feed-forward and masked self-prediction auxiliary decoders are outlined in Algorithms~\ref{algo:training-ff} and \ref{algo:training-msp} respectively.

\begin{algorithm}[H]
\caption{Training procedure for hierarchical autoregressive models. Note that training of $E_l$ and $D_l$ is carried out simultaneously in practice.}
\label{algo:training}
\begin{algorithmic}
    \State $L$: number of levels
    \State $x_l$: input representation at level $l$
    \State $x_{l+1}$: output representation at level $l$
    \State $x_1$: pixels
    \State $E_l$: encoder at level $l$
    \State $A_l$: auxiliary decoder at level $l$
    \State $D_l$: autoregressive decoder at level $l$
    \State $P$: top-level prior (at level $L$)
    \\
    \For{$l$ in $1,\ldots,L-1$}
        \State Train $E_l$, $A_l$ using auxiliary loss $\mathcal{L}_{aux}(x_l)$
        \State Train $D_l$ using conditional log-likelihood loss $\mathcal{L}_{NLL}(x_l | E_l(x_l))$
        \State $x_{l+1} := E_l(x_l)$
    \EndFor
    \\
    \State Train $P$ using log-likelihood loss $\mathcal{L}_{NLL}(x_{L})$
    \State Combine $D_1,\ldots,D_{L-1},P$ into a hierarchical autoregressive model
\end{algorithmic}
\end{algorithm}

\begin{algorithm}[H]
\caption{Training procedure for encoders with feed-forward auxiliary decoders.}
\label{algo:training-ff}
\begin{algorithmic}
    \State $x_l$: input representation at current level $l$
    \State $E_l$: encoder at level $l$
    \State $A_l$: auxiliary decoder at level $l$
    \\
    \If{l = 1}
        \State $\mathcal{L}_{aux}(x) = \sum_i \left(x_i - A_l(E_l(x_i))\right)^2$, mean-squared error in pixel space
    \Else
        \State $\mathcal{L}_{aux}(x) = -\log p_{A_l}(x | E_l(x))$, categorical negative log-likelihood in code space
    \EndIf
    \\
    \State Train $E_l$, $A_l$ using auxiliary loss $\mathcal{L}_{aux}(x_l)$
    \State Discard $A_l$, retain $E_l$
\end{algorithmic}
\end{algorithm}

\begin{algorithm}[H]
\caption{Training procedure for encoders with masked self-prediction (MSP) auxiliary decoders. Note that training of $T_l$, $E_l$ and $A_l$ is carried out simultaneously in practice.}
\label{algo:training-msp}
\begin{algorithmic}
    \State $x_l$: input representation at current level $l$
    \State $s_l$: mask size at current level $l$
    \State $m_i$: random input mask, masking one or more image regions of size $s_l \times s_l$
    \State $m_o$: output mask, masking all except the middle pixels in the masked out regions in $m_i$
    \State $E_l$: encoder at level $l$
    \State $A_l$: auxiliary decoder at level $l$
    \State $T_l$: teacher at level $l$
    \\
    \State Train $T_l$ using masked self-prediction loss $\mathcal{L}_{MSP}(x_l, s_l) = -m_o \cdot \log p_{T_l}(x_l | x_l \cdot m_i)$
    \State Train $E_l$, $A_l$ using masked distillation loss $\mathcal{L}_{dist}(x_l) = m_o \cdot D_{KL}(p_{T_l}(x_l | x_l \cdot m_i) || p_{A_l}(x_l|E_l(x_l)))$
    \State Discard $T_l$, $A_l$, retain $E_l$
\end{algorithmic}
\end{algorithm}

\section{Sampling}

The procedure for sampling from hierarchical models is outlined in Algorithm~\ref{algo:sampling}.

\begin{algorithm}[H]
\caption{Ancestral sampling procedure for hierarchical autoregressive models.}
\label{algo:sampling}
\begin{algorithmic}
    \State $L$: number of levels
    \State $x_l$: input representation at level $l$
    \State $x_1$: pixels
    \State $D_l$: autoregressive decoder at level $l$
    \State $P$: top-level prior (at level $L$)
    \\
    \State Sample $\tilde{x}_L \sim p_P(x_L)$
    \For{$l$ in $L-1,\ldots,1$}
        \State Sample $\tilde{x}_l \sim p_{D_l}(x_l|\tilde{x}_{l+1})$
    \EndFor
    \\
    \State Return $\tilde{x}_1$
\end{algorithmic}
\end{algorithm}

We report all sampling timings using a single NVIDIA V100 GPU. We use a version of incremental sampling~\cite{DBLP:journals/corr/PaineKCZRHH16} which uses buffers to avoid unnecessary recomputation. Because our current version of incremental sampling does not support models with attention, we use naive sampling for sampling for the autoregressive priors: at every point we simply pass in the entire previously sampled input to the model. For 128$\times$128 it takes roughly 23 minutes to sample a batch of 25 codes from the prior and 9 minutes to use the level 1 model to map these codes to 128$\times$128 images. For a batch of 9 images at 256$\times$256 resolution, it takes 10 minutes to first sample the level 2 codes from the prior, less than 2 minutes to sample level 1 codes conditioned on these level 2 codes, and finally 9 minutes to use these codes to sample the 256$\times$256 images themselves. 

\section{PixelCNN bias towards local structure}
\label{sec:bias-towards-local-structure}
\begin{figure}
    \centering
    \includegraphics[width=0.90\linewidth]{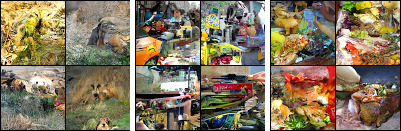}
    \caption{Samples from a gated PixelCNN trained on 64$\times$64 ImageNet images. From left to right, the classes are `lion', `ambulance' and `cheeseburger'.}
    \label{fig:baseline-samples}
\end{figure}

To demonstrate that autoregressive models like PixelCNN are inherently biased towards capturing local structure, we trained a class-conditional gated PixelCNN model~\cite{NIPS2016_6527} with 20 layers, 384 hidden units and 1024 hidden units for the residual connections on 64$\times$64 ImageNet images. Some conditional samples from this model are shown in Figure~\ref{fig:baseline-samples}. While these samples feature recognisable textures associated with the respective classes, they are not globally coherent. Generating coherent samples would require a model that is too large to train in a feasible amount of time.

\section{Effect of auxiliary decoder design on reconstructions}
\begin{figure}
  \centering
  \includegraphics[width=1.0\linewidth,trim={0cm 0cm 0cm 0cm},clip]{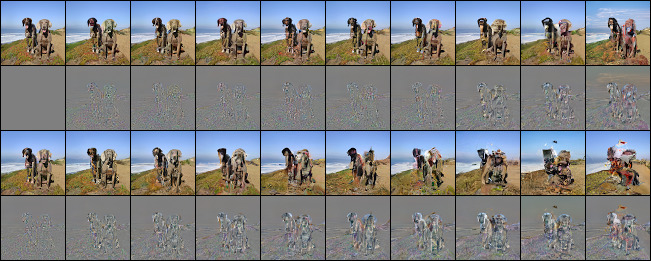}
  ~\\
  \includegraphics[width=1.0\linewidth,trim={0cm 0cm 0cm 0cm},clip]{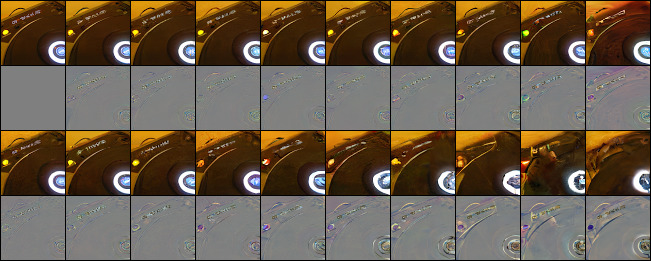}
  ~\\
  \includegraphics[width=1.0\linewidth,trim={0cm 0cm 0cm 0cm},clip]{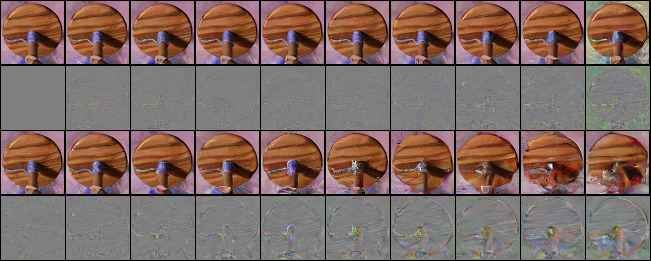}
  \caption{Reconstructions from MSP-trained codes with different auxiliary decoder hyperparameters. The top left image in each grid is the original 64$\times$64 image. The rest of row 1 shows sampled reconstructions for decreasing decoder depth (16, 14, 12, 10, 8, 6, 4, 2, 0 layers respectively) with mask size 5$\times$5. Row 3 shows reconstructions for increasing mask size (1, 3, 5, 7, 9, 11, 13, 15, 17, 19 respectively) with 2 decoder layers. Rows 2 and 4 show the difference between the rows above and the original image.}
  \label{fig:aux-decoder-design-reconstructions}
\end{figure}

In the main paper, we discussed the effect of the architecture of the auxiliary decoder on the information content and compressibility of the codes. Here, we show how this affects reconstruction quality by using the same trained models to reconstruct some example 64$\times$64 images and visualising the result. Figure~\ref{fig:aux-decoder-design-reconstructions} shows how changing the number of layers and the mask size of an MSP decoder affects sampled reconstructions. Note that the autoregressive decoder is stochastic, so repeated sampling yields slightly different reconstructions (not shown). We also show difference images to make subtle differences easier to spot. Smaller decoders lead to worse reconstruction quality overall. Larger mask sizes have the same effect and particularly affect local detail.

\section{Bounding the marginal likelihood by the joint likelihood}
As discussed in the main paper, the joint likelihood across the image pixels and all code representations can be used as a lower bound for the marginal likelihood of the image pixels only. Let $\vec{x}$ be an image, and $\vec{z}_l$ the representation of the image for each level $l$ in a hierarchy of $L$ levels (so $\vec{z}_1 = \vec{x}$). Then we can compute:

\begin{equation}
    p(\vec{x}, \vec{z}_2, \ldots, \vec{z}_L) = 
    p(\vec{z}_L) \cdot p(\vec{z}_{L-1}|\vec{z}_L) \cdot \ldots \cdot p(\vec{x}|\vec{z}_2)
\end{equation}

Because all encoders are deterministic, they associate only one set of codes with each image. If the corresponding decoders had infinite capacity, this would imply that the marginal distribution $p(\vec{x})$ is equal to the joint distribution $p(\vec{x}, \vec{z}_2, \ldots, \vec{z}_L)$. In practice however, the decoders are imperfect and they will assign some non-zero probability to an input when conditioned on other codes: let $\tilde{\vec{z}}_2 \neq \vec{z}_2$, then $p(\vec{x}|\tilde{\vec{z}}_2) > 0$. This implies that to calculate $p(\vec{x})$ correctly, we would have to integrate $p(\vec{x}, \vec{z}_2, \ldots, \vec{z}_L)$ over all codes. This is intractable, but instead we can use $p(\vec{x}, \vec{z}_2, \ldots, \vec{z}_L)$ as a lower bound for $p(\vec{x})$. As mentioned in the main text, $p(\vec{x}|\vec{z}_2)$ will dominate this bound in practice, and it is not suitable for measuring whether a model captures large-scale structure.

\section{Human evaluation}

For our human evaluation experiments, we used a varied subset of 20 classes in an attempt to cover the different types of object classes available in the ImageNet dataset, and used 50 128$\times$128 samples for each class (the same ones that we have made available online), for a total of 1,000 images per model. Where real images were used, we used the same rescaling strategy as during training (see main text). To display the images to the raters, they were first upscaled to 512$\times$512 using bilinear interpolation. We also show the ImageNet class label to the raters.

For the realism rating experiments, each image was rated 3 times by different individuals, for a total of 6,000 ratings. Raters were explicitly asked not to pay attention to scaling artifacts, but this is hard to control for, and the relatively low score for real images indicates that this has impacted the results. They were also shown examples of real and generated images, so they would have an idea of what to look out for.

For the pairwise comparison experiments, we asked the raters ``which image looks the most realistic?''. When comparing against real images, we asked instead ``which image is real?''. This nuance is important as it gives the raters some extra information. Each pair consists of one sample from the first model, and one sample from the second model. We created 500 random sample pairs per class, in such a way that every sample is used exactly 10 times (using all possible pairs would require too many ratings). Each pairing was shown in random order and rated by 3 different individuals, for a total of 30,000 ratings.

A very small fraction of answers for each experiment were unusable (we report this fraction in the `unknown' column). For the experiment comparing samples from a hierarchical model trained with an MSP auxiliary decoder versus one trained with a feed-forward auxiliary decoder, we found that all 3 raters agreed in 63.68\% of cases (accounting for 53 unusable answers, or 0.18\%). The samples from the MSP model are preferred in 50.89\% of cases, and the samples from the feed-forward model are preferred 48.93\% of cases. Although this result is very balanced, there are larger differences between the models within each class. For this experiment, we report the preferences for each class in Table~\ref{tab:human-eval-perclass}.

Our pairwise comparison experiments are similar to those conducted by \citet{DBLP:journals/corr/abs-1904-01121}, with a few key differences: our models are class-conditional, so we evaluate over a diverse subset of classes and we need to use a larger number of images. As a consequence of this, we gather fewer evaluations per image. We do not provide the raters with immediate feedback when comparing against real images: we do not tell them whether they guessed correctly or not. It is impossible to provide such feedback when comparing models directly against each other (because neither image is real), so for consistency across experiments, we do not do this even when it is technically possible.

\begin{table}
\caption{Preference for 128$\times$128 samples from hierarchical models trained with feed-forward auxiliary decoders and with MSP auxiliary decoders, for 20 classes (50 samples per model per class, 500 random pairwise comparisons by 3 raters, 1,500 answers per class in total).}
\label{tab:human-eval-perclass}
\begin{center}
\begin{small}
\begin{tabular}{rcc}
\toprule
           & \multicolumn{2}{c}{\sc{Preference}} \\
\sc{Class} & \sc{Feed-forward} & \sc{MSP} \\
\midrule
megalith (649)                      & 36.24\% & 63.62\%  \\
giant panda (388)                   & 36.53\% & 63.20\%  \\
cheeseburger (933)                  & 41.07\% & 58.80\%  \\
Geoffroy's spider monkey (381)      & 42.00\% & 57.80\%  \\
coral reef (973)                    & 43.20\% & 56.67\%  \\
schooner (780)                      & 46.07\% & 53.60\%  \\
Pomeranian (259)                    & 47.40\% & 52.60\%  \\
white stork (127)                   & 47.63\% & 51.97\%  \\
seashore (978)                      & 50.33\% & 49.67\%  \\
starfish (327)                      & 50.80\% & 49.00\%  \\
volcano (980)                       & 50.93\% & 48.93\%  \\
bookcase (453)                      & 51.13\% & 48.80\%  \\
Granny Smith (948)                  & 51.40\% & 48.47\%  \\
monarch butterfly (323)             & 51.87\% & 48.13\%  \\
yellow garden spider (72)           & 52.13\% & 47.73\%  \\
ambulance (407)                     & 52.60\% & 47.00\%  \\
frying pan (567)                    & 55.07\% & 44.73\%  \\
grey whale (147)                    & 55.33\% & 44.47\%  \\
tiger (292)                         & 56.93\% & 42.87\%  \\
Dalmatian (251)                     & 59.93\% & 39.80\%  \\

\bottomrule
\end{tabular}
\end{small}
\end{center}
\vskip -0.1in
\end{table}

\section{Nearest neighbours}
\begin{figure}
  \centering
  \includegraphics[width=0.80\linewidth,trim={0cm 0cm 0cm 0cm},clip]{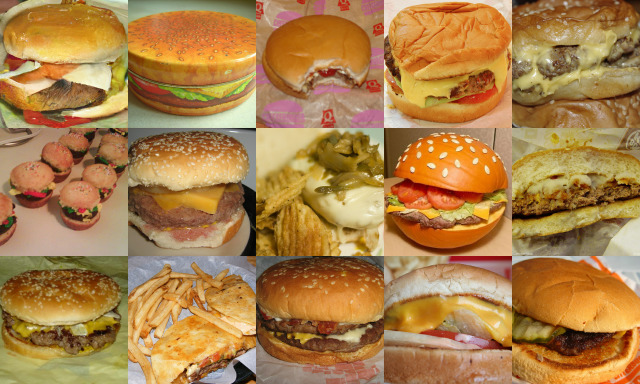}
  \caption{Nearest neighbours in pixel space. The generated image is in the top left.}
  \label{fig:nearest-neighbours-pixelspace}
\end{figure}

\begin{figure}
  \centering
  \includegraphics[width=0.80\linewidth,trim={0cm 0cm 0cm 0cm},clip]{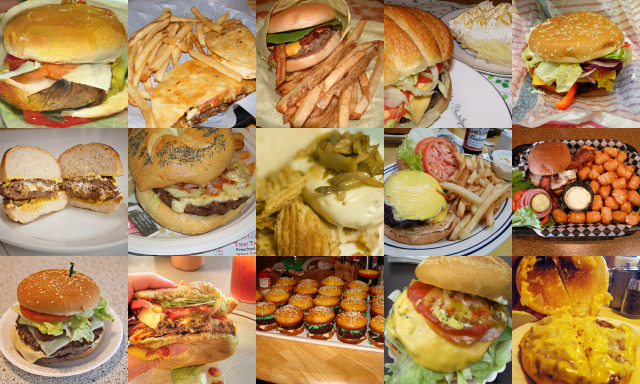}
  \caption{Nearest neighbours in VGG-16-fc7~\cite{Simonyan15} feature space. The generated image is in the top left.}
  \label{fig:nearest-neighbours-vgg}
\end{figure}

\begin{figure}
  \centering
  \includegraphics[width=0.80\linewidth,trim={0cm 0cm 0cm 0cm},clip]{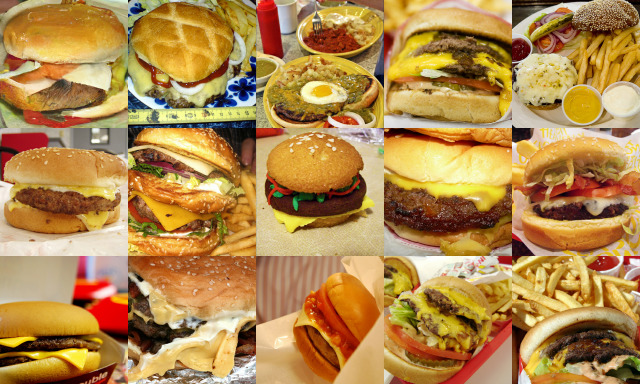}
  \caption{Nearest neighbors in ResNet-50-avgpool~\cite{He2016DeepRL} feature space. The generated image is in the top left.}
  \label{fig:nearest-neighbours-resnet}
\end{figure}

Although overfitting in likelihood-based models can be identified directly be evaluating likelihoods on a holdout set, we searched for nearest neighbours in the dataset in different feature spaces for some model samples. Following \citet{brock2018large}, we show nearest neighbours using L2 distance in pixel space, as well as in `VGG-16-fc7' and `ResNet-50-avgpool' (feature spaces obtained from pre-trained discriminative models on ImageNet) in Figures~\ref{fig:nearest-neighbours-pixelspace}, \ref{fig:nearest-neighbours-vgg} and \ref{fig:nearest-neighbours-resnet} respectively.

\section{Additional reconstructions}
\begin{figure}
  \centering
  \includegraphics[width=0.90\linewidth,trim={0cm 0cm 0cm 0cm},clip]{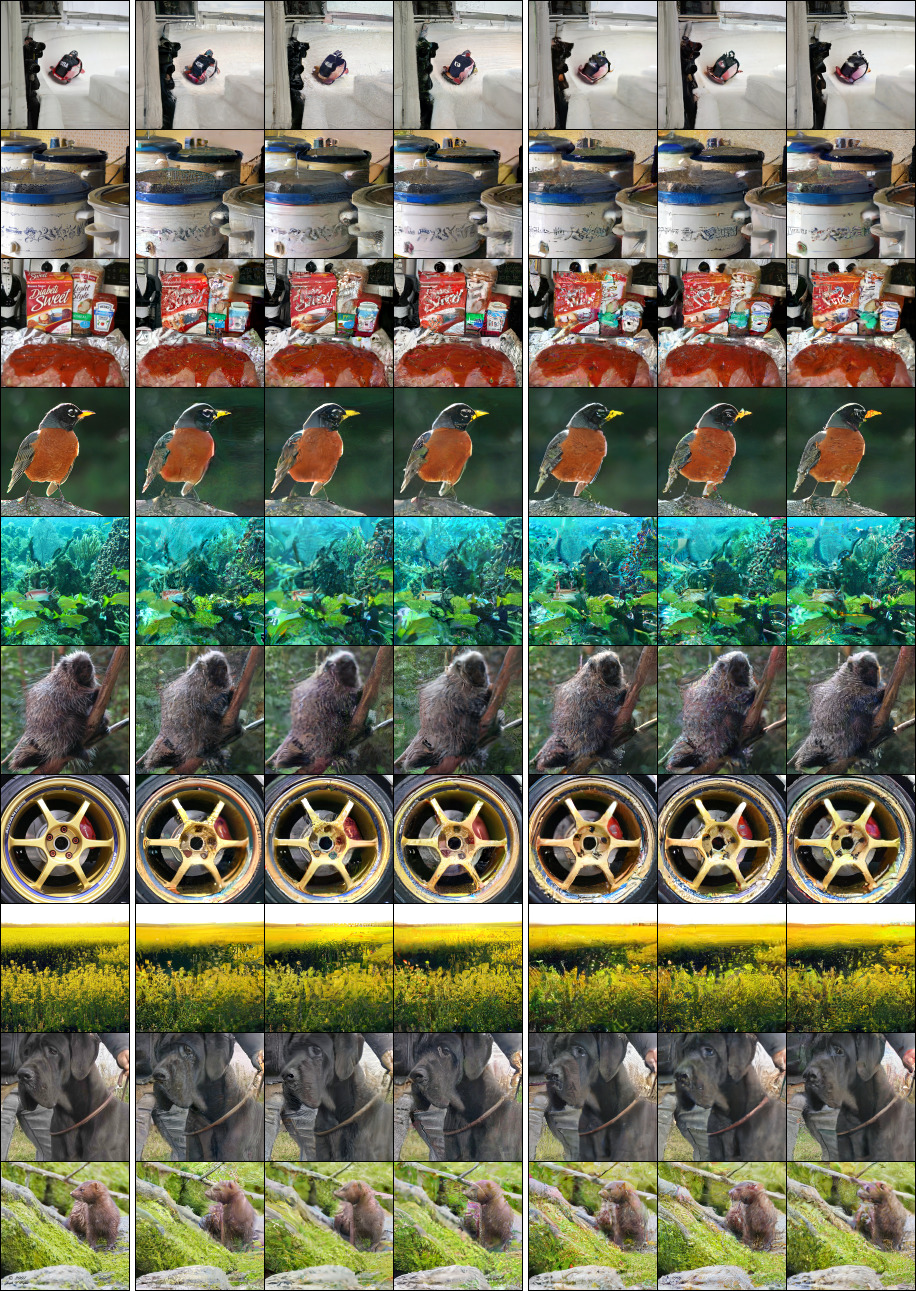}
  \caption{Additional autoregressive autoencoder reconstructions of 128$\times$128 images. Left: original images. Middle: three different sampled reconstructions from models with a feed-forward auxiliary decoder trained with the MSE loss. Right: three different sampled recontructions from models with an MSP auxiliary decoder with mask size 3$\times$3. The sampling temperature was $0.99$.}
  \label{fig:reconstructions-128x128-big}
\end{figure}

\begin{figure}
  \centering
  \includegraphics[width=1.0\linewidth,trim={0cm 0cm 0cm 0cm},clip]{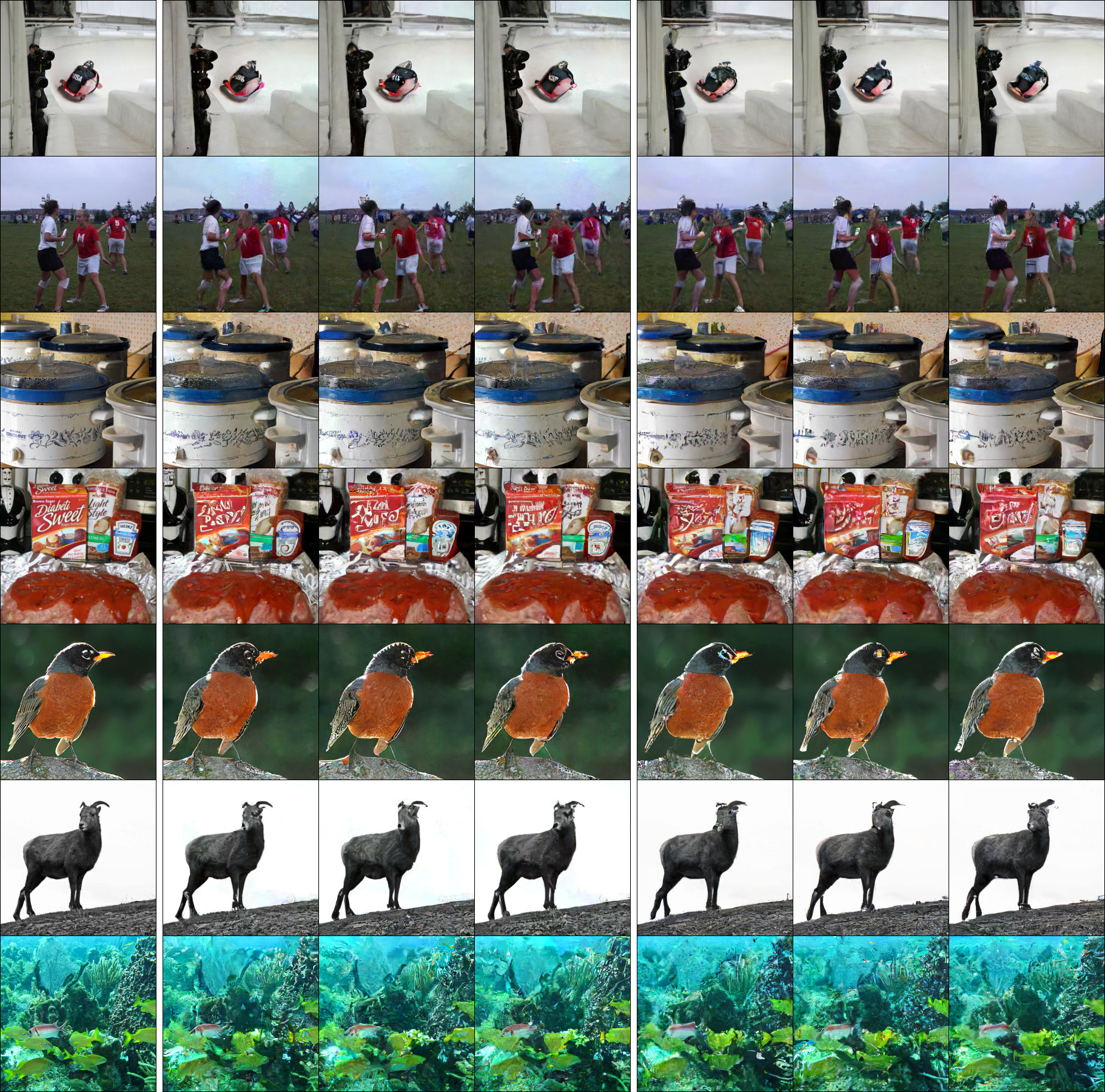}
  \caption{Additional autoregressive autoencoder reconstructions of 256$\times$256 images. Left: original images. Middle: three different sampled reconstructions from models with a feed-forward auxiliary decoder trained with the MSE loss. Right: three different sampled recontructions from models with an MSP auxiliary decoder with mask sizes 5$\times$5 (level 1) and 3$\times$3 (level 2). The sampling temperature was $0.99$.}
  \label{fig:reconstructions-256x256-big}
\end{figure}

\begin{figure}
  \centering
  \includegraphics[width=0.98\linewidth,trim={0cm 0cm 0cm 0cm},clip]{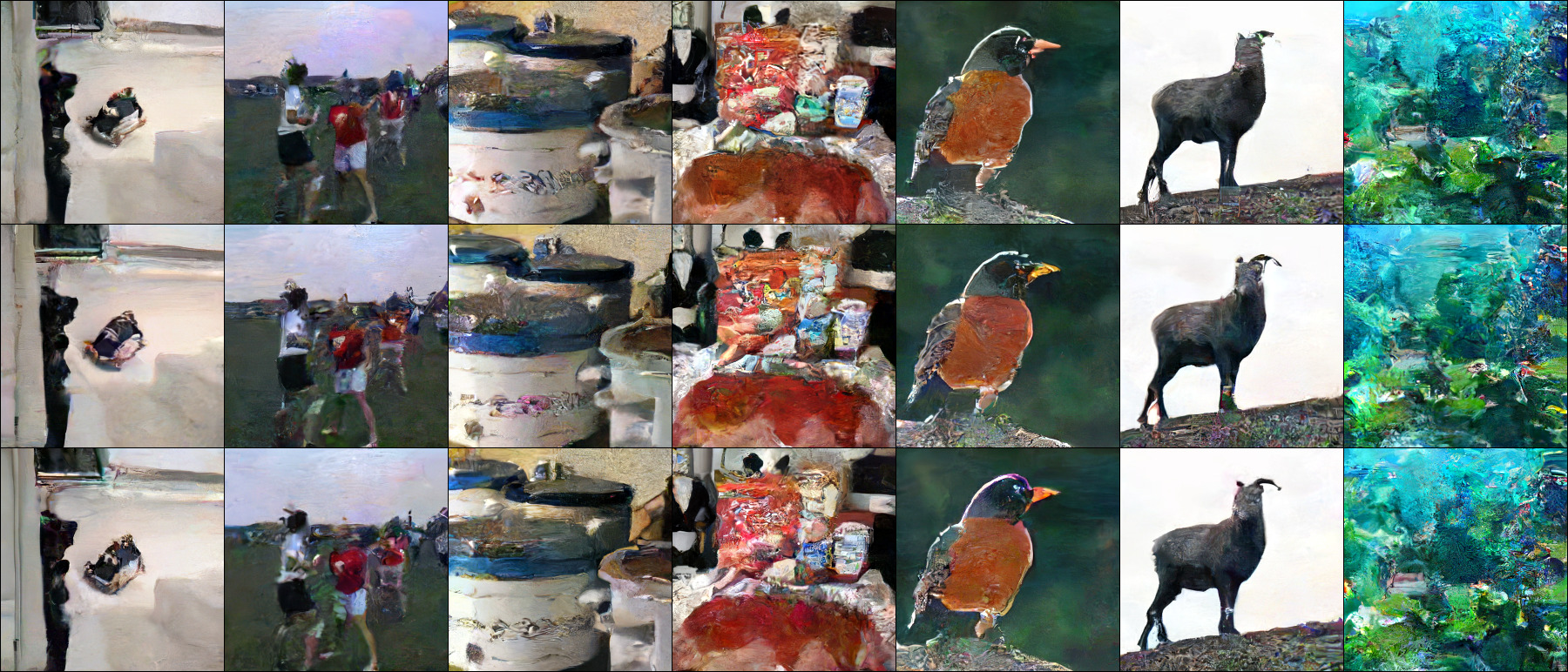}
  \caption{Autoregressive autoencoder reconstructions of 256$\times$256 images, using a single autoencoder (MSP auxiliary decoder, mask size 5$\times$5). The reconstructed images are significantly degraded in terms of visual fidelity. The sampling temperature was $0.99$. Please refer to Figure~\ref{fig:reconstructions-128x128-big} for the original images and a comparison with 2-level reconstructions.}
  \label{fig:reconstructions-256x256-direct}
\end{figure}

We provide some additional autoregressive autoencoder reconstructions of 128$\times$128 images in Figure~\ref{fig:reconstructions-128x128-big} to further demonstrate their variability and the differences between both auxiliary decoder strategies. Figure~\ref{fig:reconstructions-256x256-big} contains additional reconstructions of 256$\times$256 images. Figure~\ref{fig:reconstructions-256x256-direct} shows reconstructions from an autoencoder that directly compresses 256$\times$256 images to single-channel 32$\times$32 8-bit codes (MSP, 8 auxiliary decoder layers, mask size 5$\times$5), resulting in a significant degradation in visual fidelity.

\section{Additional samples}

\begin{figure}
  \centering
  \includegraphics[width=0.6\linewidth,trim={0cm 0cm 0cm 0cm},clip]{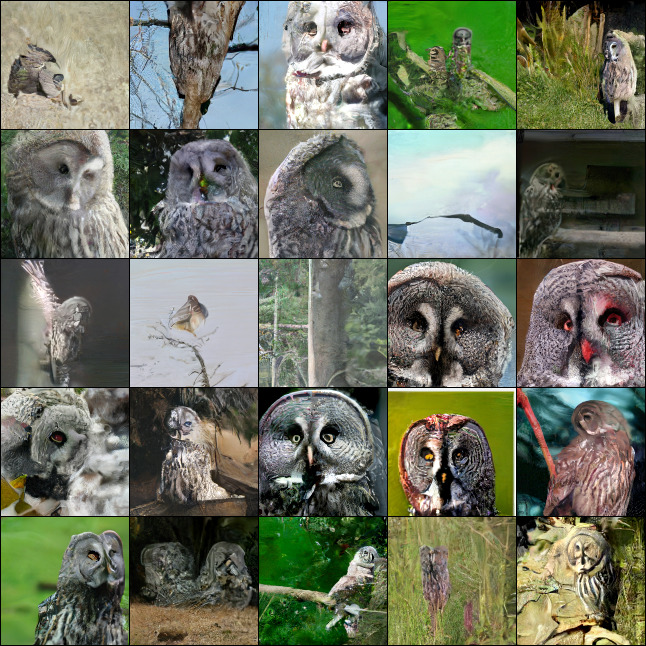}
  \caption{Great grey owl.}
  \label{fig:ff_128_first}
\end{figure}

\begin{figure}
  \centering
  \includegraphics[width=0.6\linewidth,trim={0cm 0cm 0cm 0cm},clip]{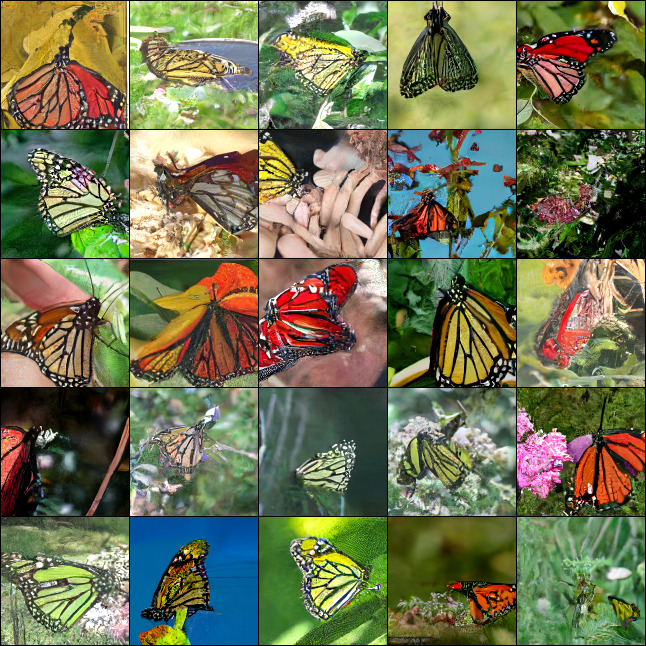}
  \caption{Monarch butterfly.}
  \label{fig:samples0}
\end{figure}

\begin{figure}
  \centering
  \includegraphics[width=0.6\linewidth,trim={0cm 0cm 0cm 0cm},clip]{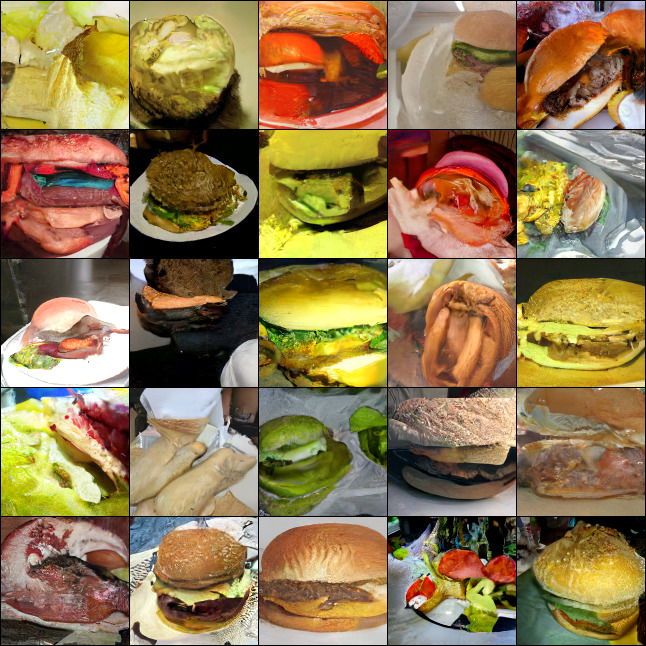}
  \vskip -0.15in
  \caption{Cheeseburger.}
  \label{fig:ff_128_last}
  \vskip -0.15in
\end{figure}

\begin{figure}
  \centering
  \includegraphics[width=0.6\linewidth,trim={0cm 0cm 0cm 0cm},clip]{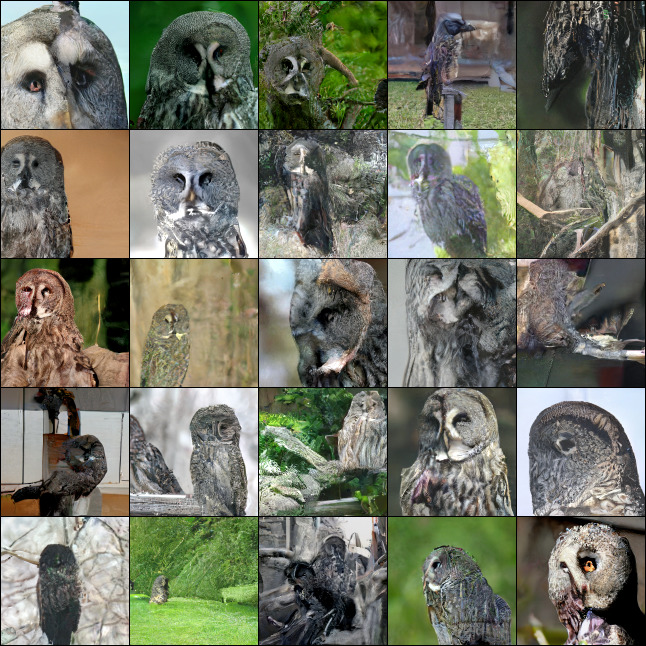}
  \caption{Great grey owl.}
  \label{fig:msp_128_first}
\end{figure}

\begin{figure}
  \centering
  \includegraphics[width=0.6\linewidth,trim={0cm 0cm 0cm 0cm},clip]{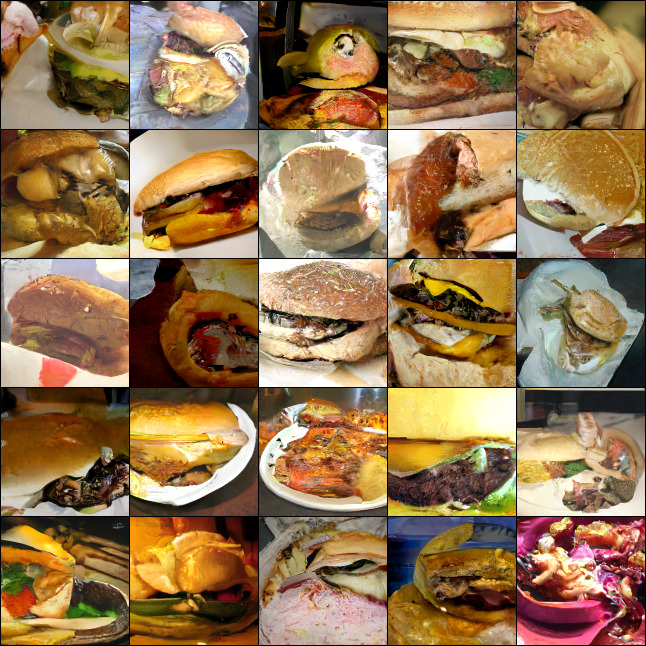}
  \caption{Cheeseburger.}
  \label{fig:msp_128_last}
\end{figure}

\begin{figure}
  \centering
  \includegraphics[width=0.6\linewidth,trim={0cm 0cm 0cm 0cm},clip]{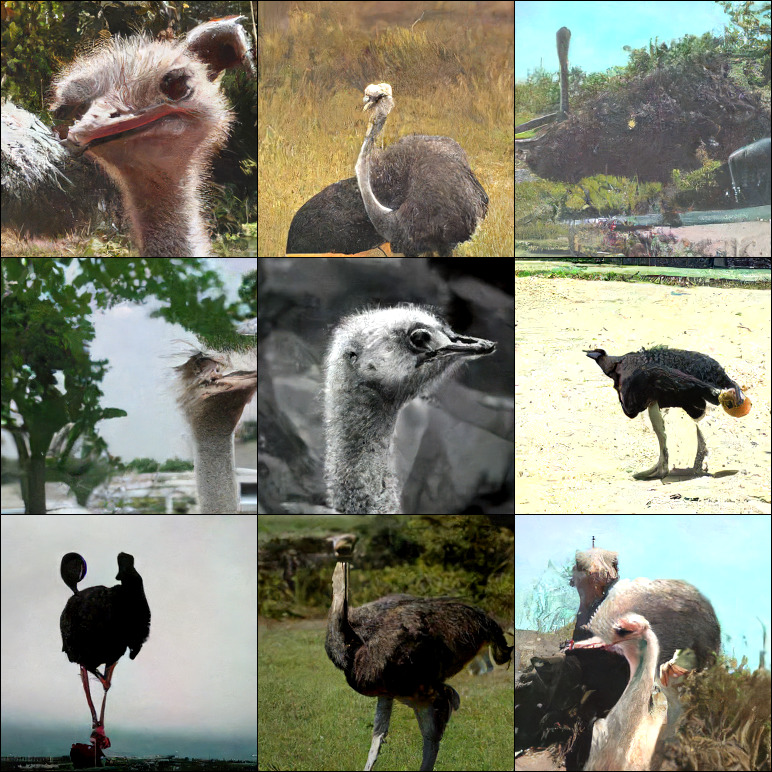}
  \caption{Ostrich.}
  \label{fig:ff_256_first}
\end{figure}

\begin{figure}
  \centering
  \includegraphics[width=0.6\linewidth,trim={0cm 0cm 0cm 0cm},clip]{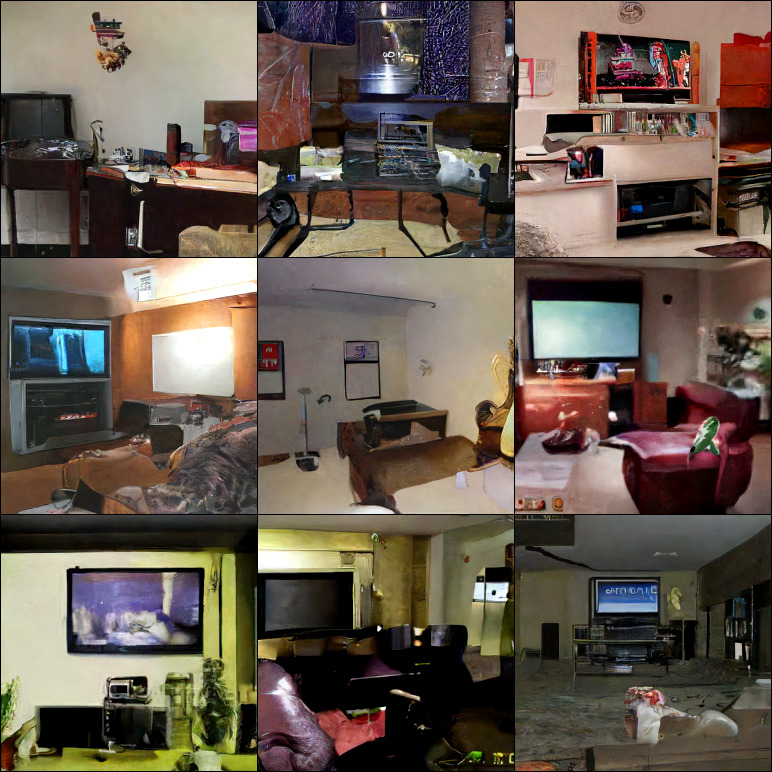}
  \caption{Home theater.}
  \label{fig:ff_256_last}
\end{figure}

\begin{figure}
  \centering
  \includegraphics[width=0.6\linewidth,trim={0cm 0cm 0cm 0cm},clip]{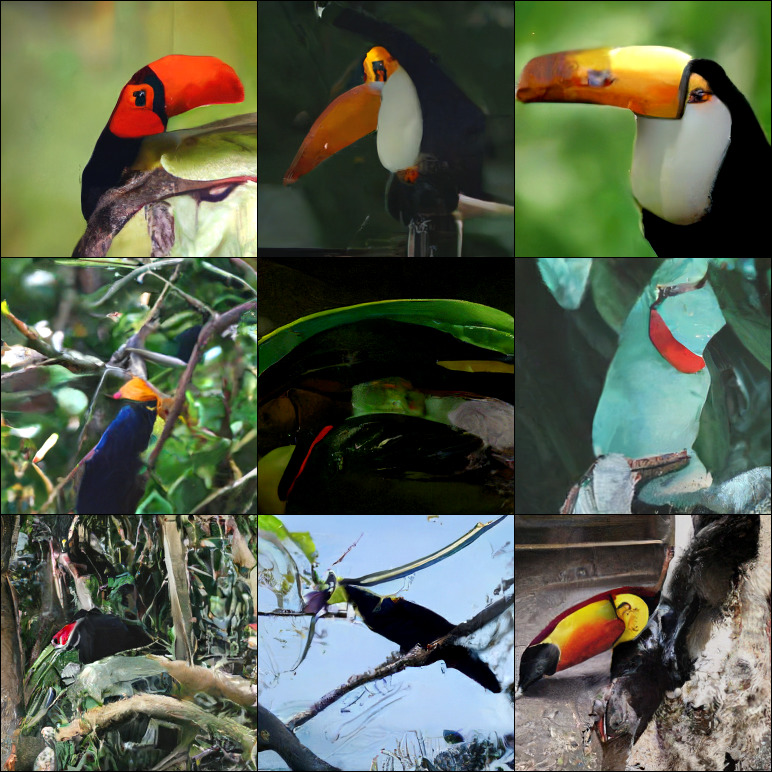}
  \caption{Toucan.}
  \label{fig:msp_256_first}
\end{figure}

\begin{figure}
  \centering
  \includegraphics[width=0.6\linewidth,trim={0cm 0cm 0cm 0cm},clip]{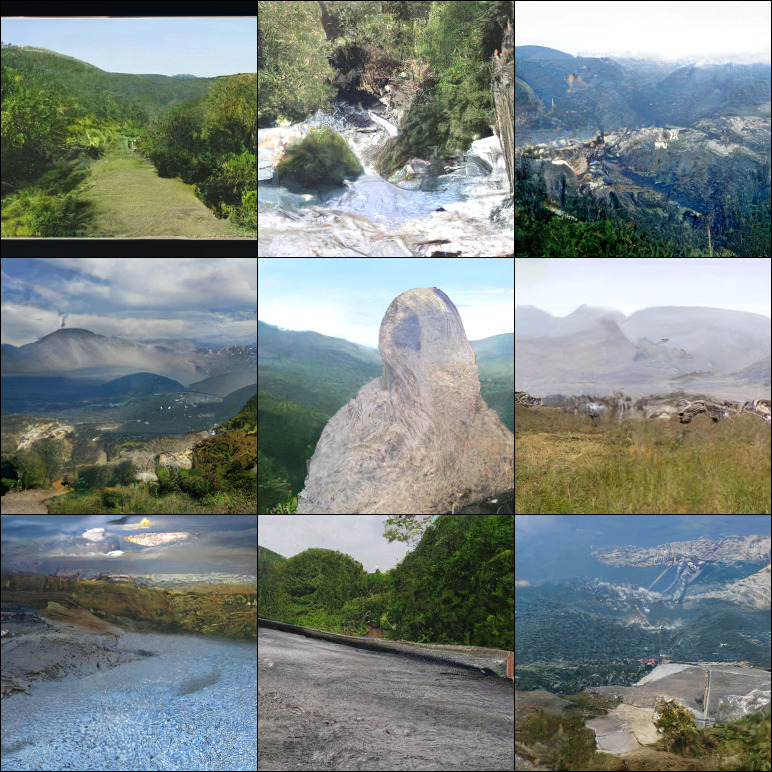}
  \caption{Valley.}
  \label{fig:msp_256_last}
\end{figure}

More samples are available online at \url{https://bit.ly/2FJkvhJ} in original quality (some figures in the paper are compressed to save space). Figures \ref{fig:ff_128_first}-\ref{fig:ff_128_last} are 128$\times$128 samples for a model using a feed-forward auxiliary decoder. Figures \ref{fig:msp_128_first}-\ref{fig:msp_128_last} are 128$\times$128 samples for a model using a MSP auxiliary decoder. Figures \ref{fig:ff_256_first}-\ref{fig:ff_256_last} are 256$\times$256 samples from a model using feed-forward auxiliary decoders. Figures \ref{fig:msp_256_first}-\ref{fig:msp_256_last} are 256$\times$256 samples from a model using MSP auxiliary decoders.


\begin{figure}
  \centering
  \includegraphics[width=0.6\linewidth,trim={0cm 0cm 0cm 0cm},clip]{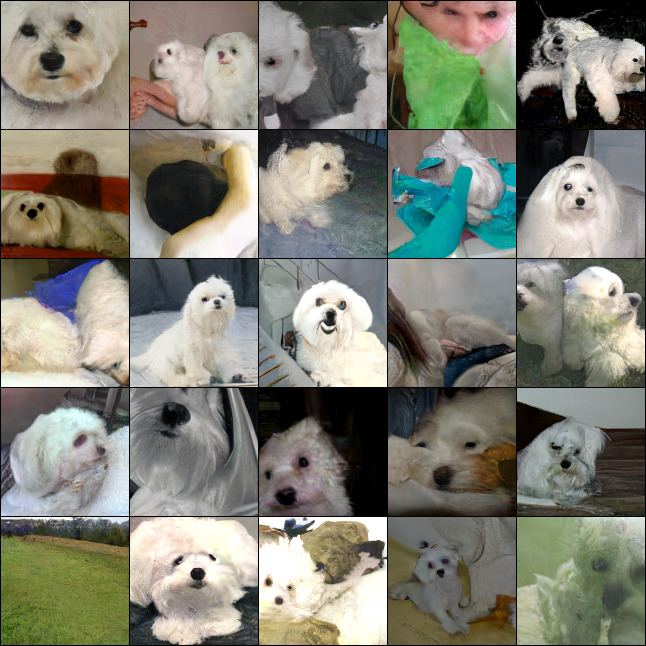}
  \caption{Maltese.}
\end{figure}


\begin{figure}
  \centering
  \includegraphics[width=0.6\linewidth,trim={0cm 0cm 0cm 0cm},clip]{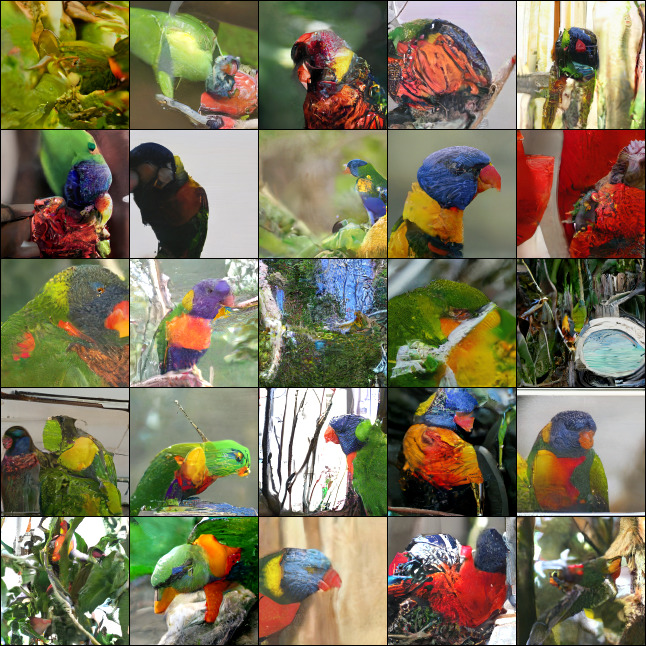}
  \caption{Lorikeet.}
\end{figure}


\begin{figure}
  \centering
  \includegraphics[width=0.6\linewidth,trim={0cm 0cm 0cm 0cm},clip]{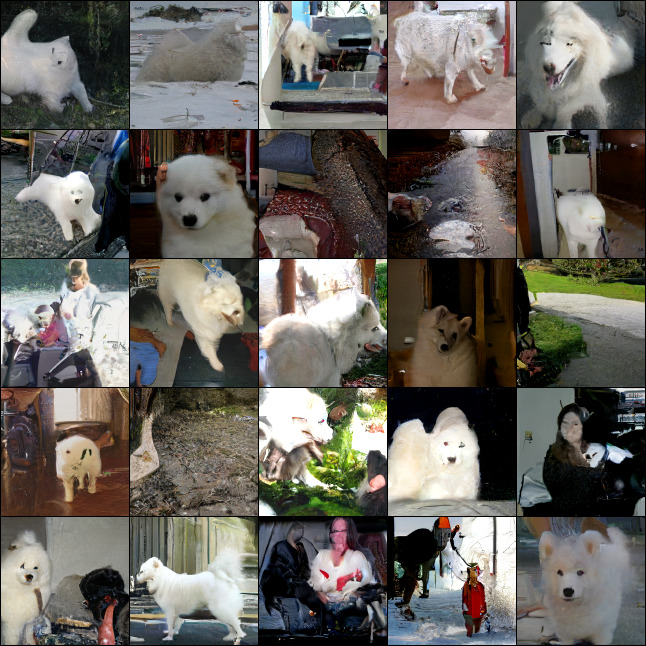}
  \caption{Samoyed.}
\end{figure}



\begin{figure}
  \centering
  \includegraphics[width=0.6\linewidth,trim={0cm 0cm 0cm 0cm},clip]{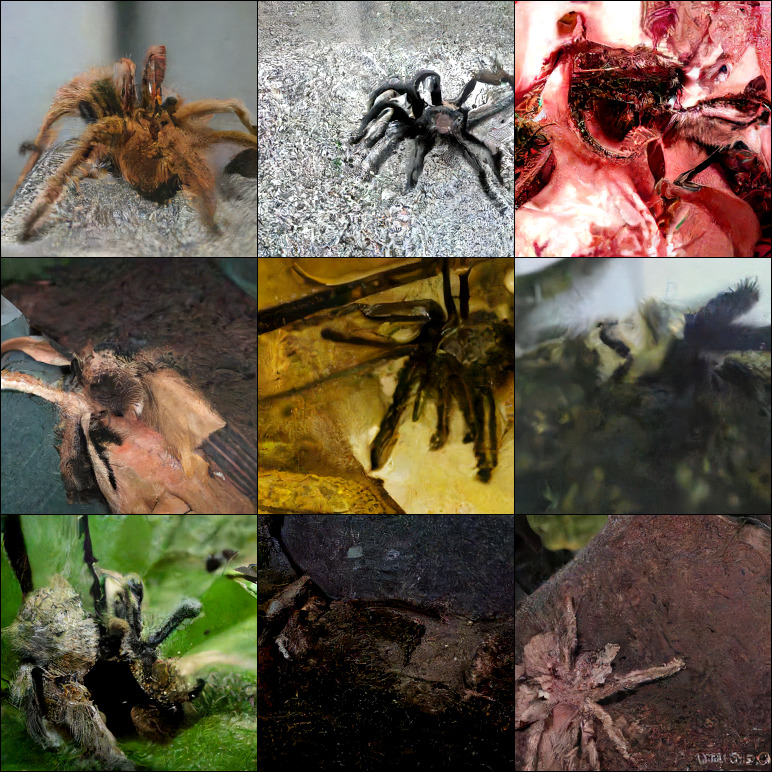}
  \caption{Tarantula.}
\end{figure}



\begin{figure}
  \centering
  \includegraphics[width=0.6\linewidth,trim={0cm 0cm 0cm 0cm},clip]{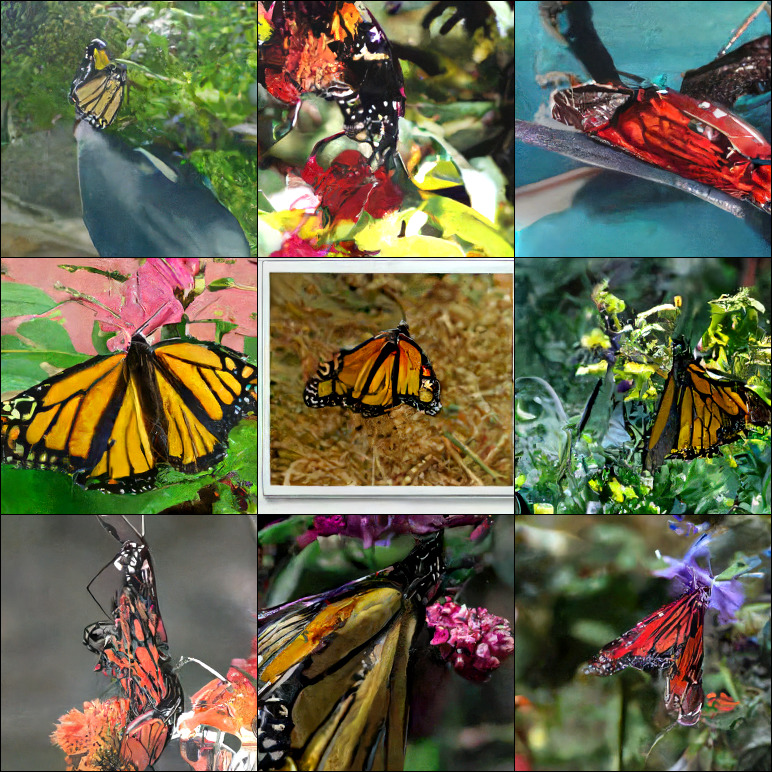}
  \caption{Monarch butterfly.}
\end{figure}

\begin{figure}
  \centering
  \includegraphics[width=0.6\linewidth,trim={0cm 0cm 0cm 0cm},clip]{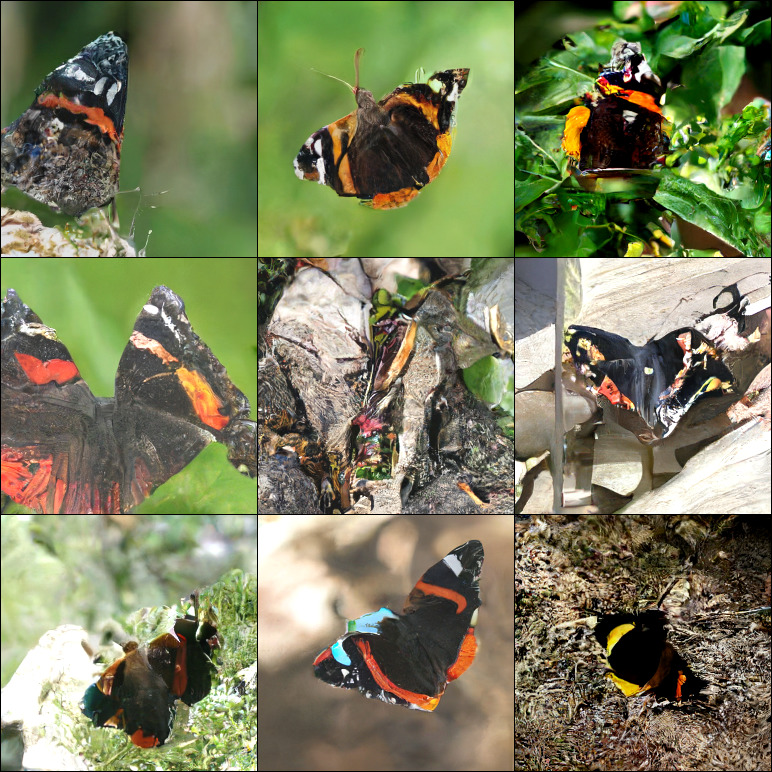}
  \caption{Red admiral.}
\end{figure}

\end{document}